\documentclass[runningheads]{llncs}

\usepackage{eccv}

\usepackage{eccvabbrv}

\usepackage{graphicx}
\usepackage{wrapfig} 
\usepackage{booktabs}

\usepackage[symbol]{footmisc}
\usepackage{tikz}
\usepackage{comment}
\usepackage{amsmath,amssymb} 
\usepackage{color}
\usepackage[accsupp]{axessibility}  
\usepackage{multicol}
\usepackage{lipsum}

\usepackage[width=122mm,left=12mm,paperwidth=146mm,height=193mm,top=12mm,paperheight=217mm]{geometry}

\usepackage{algorithm}
\usepackage{algorithmicx, algpseudocode}

\usepackage{xcolor}
\usepackage{colortbl}

\usepackage[pagebackref,breaklinks,colorlinks,citecolor=eccvblue]{hyperref}

\usepackage{orcidlink}
\usepackage{algorithm}
\usepackage{algorithmicx, algpseudocode}

\begin{document}

\title{Deblurring 3D Gaussian Splatting} 


\newcommand\CoAuthorMark{\footnotemark[\arabic{footnote}]}
\newcommand\CorrespondingAuthorMark{\footnotemark[\arabic{footnote}]}
\author{Byeonghyeon Lee\inst{1}\thanks{Equal contribution}\orcidlink{0009-0007-9911-1948} \and
Howoong Lee\inst{1,3}\protect\CoAuthorMark\orcidlink{0009-0003-3337-4914} \and
Xiangyu Sun\inst{1}\orcidlink{0009-0009-0625-4240} \and
Usman Ali\inst{1}\orcidlink{0000-0002-8986-3173} \and
Eunbyung Park\inst{1,2}\thanks{Corresponding authors}\orcidlink{0000-0003-4071-2814}
}

\authorrunning{B. Lee et al.}

\institute{Department of Electrical and Computer Engineering, Sungkyunkwan University  \and
Department of Artificial Intelligence, Sungkyunkwan University \and
Hanhwa Vision}

\maketitle

\begin{abstract}
Recent studies in Radiance Fields have paved the robust way for novel view synthesis with their photorealistic rendering quality. Nevertheless, they usually employ neural networks and volumetric rendering, which are costly to train and impede their broad use in various real-time applications due to the lengthy rendering time. 
Lately 3D Gaussians splatting-based approach has been proposed to model the 3D scene, and it achieves remarkable visual quality while rendering the images in real-time. However, it suffers from severe degradation in the rendering quality if the training images are blurry. Blurriness commonly occurs due to the lens defocusing, object motion, and camera shake, and it inevitably intervenes in clean image acquisition. Several previous studies have attempted to render clean and sharp images from blurry input images using neural fields. The majority of those works, however, are designed only for volumetric rendering-based neural radiance fields and are not straightforwardly applicable to rasterization-based 3D Gaussian splatting methods.
Thus, we propose a novel real-time deblurring framework, Deblurring 3D Gaussian Splatting, using a small Multi-Layer Perceptron (MLP) that manipulates the covariance of each 3D Gaussian to model the scene blurriness. While Deblurring 3D Gaussian Splatting can still enjoy real-time rendering, it can reconstruct fine and sharp details from blurry images. A variety of experiments have been conducted on the benchmark, and the results have revealed the effectiveness of our approach for deblurring. Qualitative  results are available at \href{https://benhenryl.github.io/Deblurring-3D-Gaussian-Splatting/}{https://benhenryl.github.io/Deblurring-3D-Gaussian-Splatting/}
\end{abstract}

\section{Introduction}
\label{sec:intro}

With the emergence of Neural Radiance Fields (NeRF)~\cite{nerf}, Novel view synthesis (NVS) has accounted for more roles in computer vision and graphics with its photorealistic scene reconstruction and applicability to diverse domains such as augmented/virtual reality (AR/VR) and robotics.
Various NVS methods typically involve modeling 3D scenes from multiple 2D images from arbitrary viewpoints, and these images are often taken under diverse conditions. One of the significant challenges, particularly in practical scenarios, is the common occurrence of blurring effects. It has been a major bottleneck in rendering clean and high-fidelity novel view images, as it requires accurately reconstructing the 3D scene from the blurred input images.

NeRF~\cite{nerf} has shown outstanding performance in synthesizing photo-realistic images for novel viewpoints by representing 3D scenes with implicit functions. The volume rendering~\cite{drebin1988volume} technique has been a critical component of the massive success of NeRF. This can be attributed to its continuous nature and differentiability, making it well-suited to today's prevalent automatic differentiation software ecosystems. However, significant rendering and training costs are associated with the volumetric rendering approach due to its reliance on dense sampling along the ray to generate a pixel, which requires substantial computational resources. Despite the recent advancements~\cite{garbin2021fastnerf,sun2022direct,muller2022instant,kplanes_2023,fridovich2022plenoxels} that significantly reduce training time from days to minutes, improving the rendering time still remains a vital challenge.

\begin{figure}[t]
\begin{center}
\includegraphics[width=4.8in, keepaspectratio]{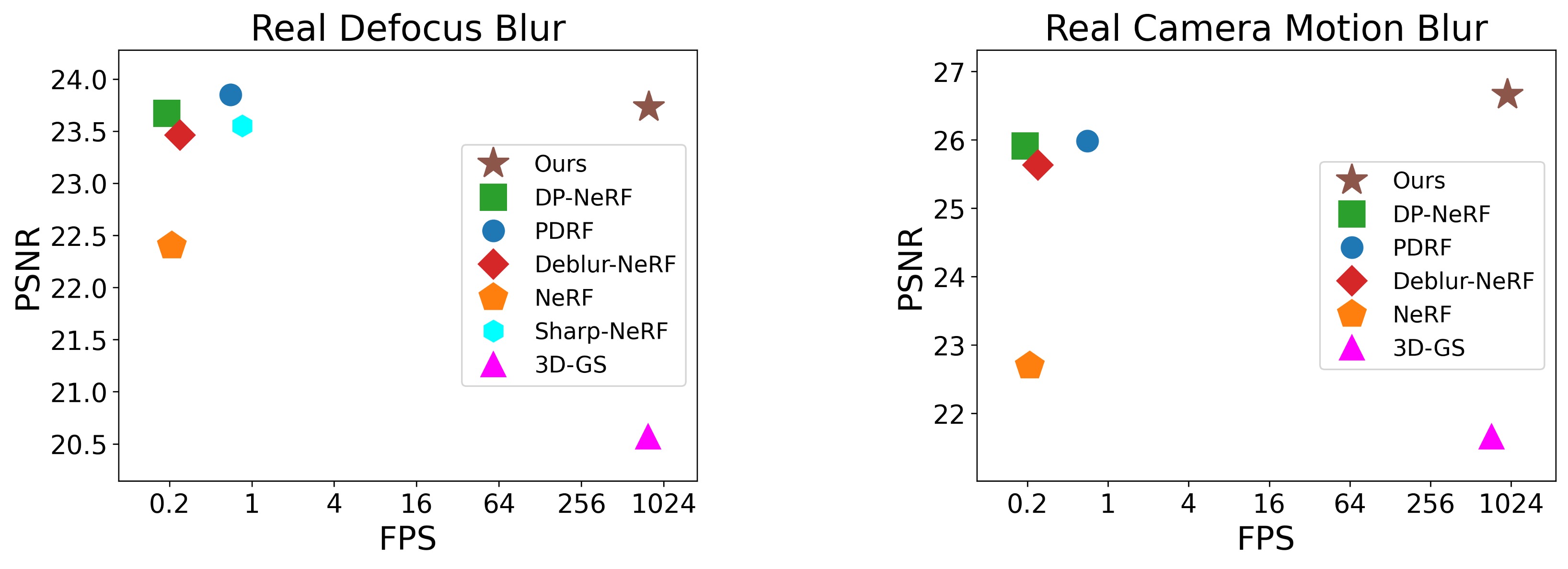}
\end{center}
   \caption{Performance comparison to state-of-the-art deblurring NeRFs. Ours achieved a fast rendering speed ($>$ 800 FPS vs. 1 FPS) while maintaining competitive rendered image quality (the x-axis is represented in log scale).}
\label{fig:curve}
\end{figure}


Recently, 3D Gaussian Splatting (3D-GS)~\cite{kerbl20233d} has gained significant attention, demonstrating a capability to produce high-quality images at a remarkably fast rendering speed. 
Substituting NeRF's time-demanding volumetric rendering, it combines a large number of colored 3D Gaussians to represent 3D scenes with a differentiable splatting-based rasterization, which can be significantly more efficient than volume rendering techniques on modern graphics hardware, thereby enabling rapid real-time rendering.

Expanding on the impressive capabilities of 3D-GS, we aim to further improve its robustness and versatility for more realistic settings, especially those involving blurring effects. Several approaches have attempted to handle the blurring issues in the recent NeRF literature~\cite{ma2022deblur,Lee_2023_CVPR,dai2023hybrid,wang2022bad,dof-nerf}.
The pioneering work is Deblur-NeRF~\cite{ma2022deblur}, which renders sharp images from images with defocus blur or camera motion blur using an extra multi-layer perceptron (MLP) to produce the blur kernels.
DP-NeRF~\cite{Lee_2023_CVPR} constrains neural radiance fields with two physical priors derived from the actual blurring process to reconstruct clean images. PDRF~\cite{peng2022pdrf} uses a two-stage deblurring scheme and a voxel representation to further improve deblurring and training time. All works mentioned above have been developed under the assumption of volumetric rendering, which is not straightforwardly applicable to rasterization-based 3D-GS. Another line of works~\cite{wang2022bad,dai2023hybrid,sharpnerf} though not dependent on volume rendering, only address a single specific type of blur, i.e., either camera motion blur or defocus blur, and are not valid for mitigating the both types of blur.

In this work, we propose Deblurring 3D-GS, the first deblurring algorithm for 3D-GS, which is well aligned with rasterization and thus enables real-time rendering. To do so, we modify the covariance matrices of 3D Gaussians to model the blurriness. Specifically, we employ a small MLP, which manipulates the covariance mean of each 3D Gaussian to model the scene blurriness. 
As blurriness is a phenomenon that is based on the intermingling of the neighboring pixels, our Deblurring 3D-GS simulates such an intermixing during the training time. To this end, we designed a framework that utilizes an MLP to learn the variations in different attributes of 3D Gaussians. These small variations are multiplied or added to the original values of the attributes, which in turn determine the updated shape of the resulting Gaussians. During the inference time, we render the scene using only the original components of 3D-GS without any additional outputs from the MLP; thereby, 3D-GS can render sharp images because each pixel is free from the intermingling of nearby pixels. Further, since the MLP is not activated during the inference time, it can still enjoy real-time rendering similar to the 3D-GS while it can reconstruct fine and sharp details from the blurry images. 

\begin{figure*}[t!]
    \centering
    \includegraphics[width=4.8in, keepaspectratio]{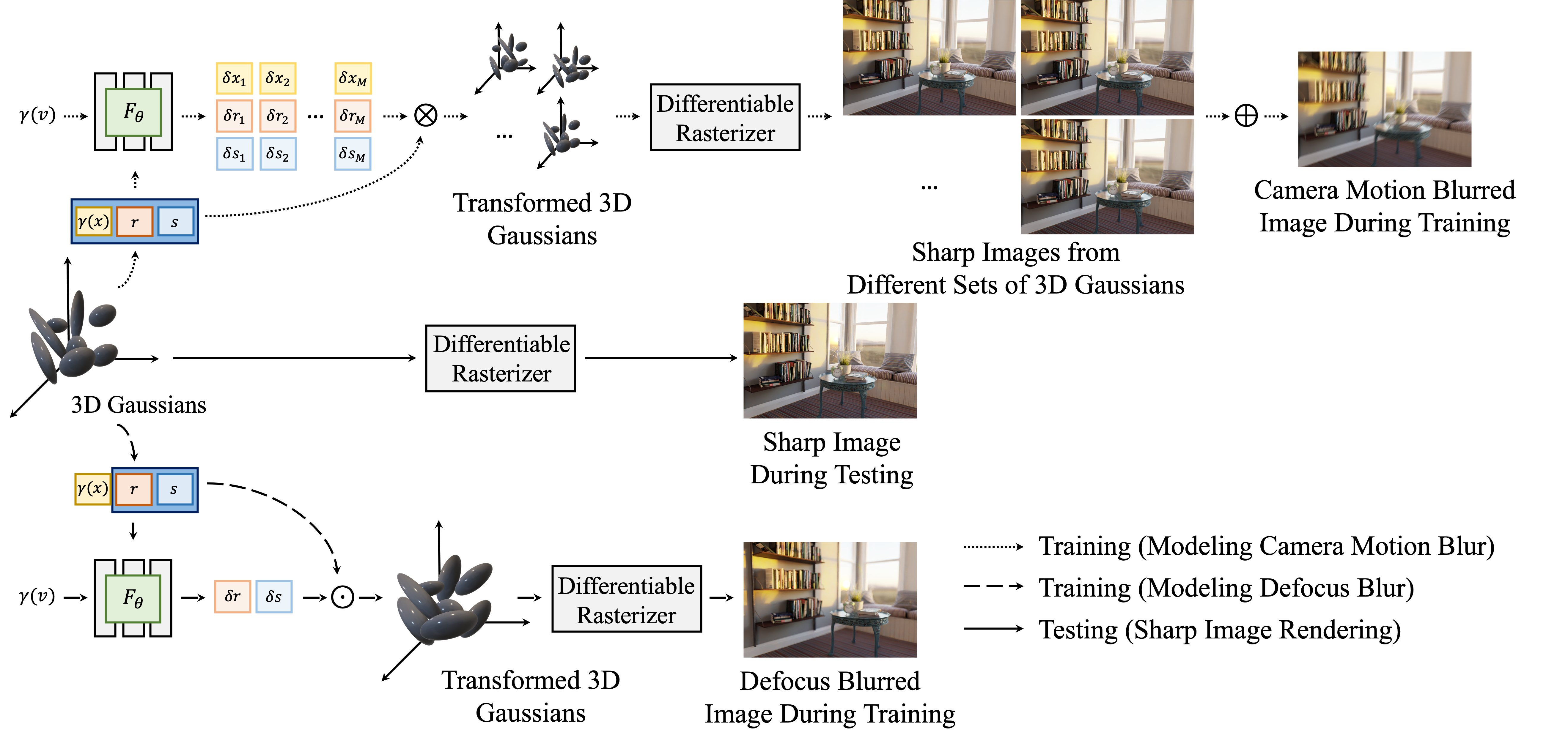}
    \caption{Our method's overall workflow. $\gamma(\cdot)$ denotes positional encoding, $\odot$ and $\oplus$ denotes hadamard product and averaging operation for each, and $x$, $r$, $s$ stand for position, quaternion, and scaling of 3D Gaussian respectively. $\otimes$ is an operator that implements $\delta r \odot r$, $\delta s \odot s$, and $\delta x + x$. Dotted arrows and dashed arrows describe the pipeline for modeling camera motion blur and modeling defocus blur, respectively at training time. Solid arrows show the process of rendering sharp images at the inference time. More details are explained at ~\cref{sec:deblurring_3dgs}.
    }
    \label{fig:workflow}
\end{figure*}

3D-GS~\cite{kerbl20233d} models a 3D scene from a sparse point cloud, which is usually obtained from the structure-from-motion (SfM)~\cite{schonberger2016structure}. SfM extracts features from multi-view images and relates them via 3D points in the scene. If the given images are blurry, SfM fails heavily in identifying the valid features, and ends up extracting a very small number of points. Even worse, if the scene has a larger depth of field, SfM hardly extracts any points which lie on the far end of the scene. Due to this excessive sparsity in the point cloud constructed from set of blurry images, existing methods, including 3D-GS~\cite{kerbl20233d}, that rely on point clouds fail to reconstruct the scene with fine details. To compensate for this excessive sparsity, we propose to add extra points with valid color features to the point cloud using K-nearest-neighbor interpolation{~\cite{knn}}. In addition, we prune Gaussians based on their position to keep more Gaussians on the far plane.

A variety of experiments have been conducted on the benchmark, and the results have revealed the effectiveness of our approach for deblurring. Tested under different evaluation matrices, our method achieves state-of-the-art rendering quality or performs on par with the currently leading models while achieving significantly faster rendering speed ($>$ 800 FPS)

To sum up, our contributions are the following:
\begin{itemize}
    \item We propose the first real-time rendering deblurring framework using 3D-GS.
    \item We propose a novel technique that manipulates the covariance matrix and mean of each 3D Gaussian differently to model spatially changing blur using a small MLP.
    \item To compensate for sparse point clouds due to the blurry images, we propose a training technique that prunes and adds extra points with valid color features so that we can put more points on the far plane of the scene and harshly blurry regions.
    \item We achieve FPS $>$ 800 while accomplishing superior rendering quality or performing on par with the existing cutting-edge models under different metrics.
\end{itemize}

\section{Related Works}
\label{sec:related_works}

\subsection{Image Deblurring}
It is common to observe that when we casually take pictures with optical imaging systems, some parts of the scene appear blurred in the images. This blurriness is caused by a variety of factors, including object motion, camera shake, and lens defocusing~\cite{abuolaim2022improving,Ruan2022learning}.
The degradation induced by the blur of an image is generally expressed as follows: 

\begin{equation}
g(x) = \sum_{s\in S_h} h(x,s)f(x) + n(x), ~x \in S_f,
\end{equation}
\label{Eq:blur_convolution}
where $g(x)$ represents an observed blurry image, $h(x,s)$ is a blur kernel or Point Spread Function (PSF), $f(x)$ is a latent sharp image, and $n(x)$ denotes an additive white Gaussian noise that frequently occurs in nature images. $S_f \subset \mathbb{R}^2$ is a support set of an image and $S_h \subset \mathbb{R}^2$ is a support set of a blur kernel or PSF~\cite{kundur1996blind}.

Traditional methods often construct deblurring as an optimization problem and rely on natural image priors~\cite{liu2020estimating,pan2016blind,xu2013unnatural,zhang2022pixel}. Conversely, the majority of deep learning-based techniques use convolutional neural networks (CNN) to map the blurry image with the latent sharp image directly~\cite{nimisha2017blur,zhang2019deep,ren2020neural}. While a series of studies have been actively conducted for image deblurring, they are mainly designed for deblurring 2D images and are not easily applicable to 3D scenes deblurring due to the lack of 3D view consistency. 

\subsection{Neural Radiance Fields}
Neural Radiance Fields (NeRF) is a potent method that has gained popularity for creating high-fidelity 3D scenes from 2D images, employing deep neural networks to encode volumetric scene features. To estimate density $\sigma \in [0,\infty)$ and color value $c \in [0, 1]^3$ of a given point, a radiance field is a continuous function $f$ that maps a 3D location $x \in \mathbb{R}^3$ and a viewing direction $d \in \mathbb{S}^2$.  This function has been parameterized by a multi-layer perceptron (MLP)~\cite{nerf}, where the weights of MLP are optimized to reconstruct a series of input photos of a particular scene: $(c, \sigma) = f_\theta : \big(\gamma(x), \gamma(d) \big)$. Here, $\theta$ indicates the network weights, and $\gamma$ is the specified positional encoding applied to $x$ and $d$~\cite{tancik2020fourier}. To generate the images at novel views, volume rendering~\cite{drebin1988volume} is used, taking into account the volume density and color of points. 

\subsubsection{Fast Inference NeRF} Numerous follow-up studies have been carried out to enhance NeRF's rendering time to achieve real-time rendering. Many methods, such as grid-based approaches~\cite{f2nerf,chen2022tensorf,sun2022direct,Hexplane,kplanes_2023,Rho_2023_CVPR,mipgrid}, or those relying on hash~\cite{muller2022instant,barron2023zipnerf} adopt additional data structures to effectively reduce the size and number of layers of MLP and successfully improve the inference speed. However, they still fail to reach real-time view synthesis. Another line of works~\cite{Reiser2023MERF,hedman2021snerg,yariv2023bakedsdf} proposes to bake the trained parameters into the faster representation and attain real-time rendering. While these methods rely on volumetric rendering, recently, 3D-GS~\cite{kerbl20233d} successfully renders photo-realistic images at novel views with noticeable rendering speed using a differentiable rasterizer and 3D Gaussians. Although several approaches have attempted to render tens or hundreds of images in a second, deblurring the blurry scene in real-time is not addressed, while blurriness commonly hinders clean image acquisition in the wild.

\subsubsection{Deblurring NeRF} 
Several strategies have been proposed to train NeRF to render clean and sharp images from blurry input images. While DoF-NeRF~\cite{dof-nerf} attempts to deblur the blurry scene, both all-in-focus and blurry images are required to train the model. Deblur-NeRF~\cite{ma2022deblur} firstly suggests deblurring NeRF without any all-in-focus images during training. It employs an additional small MLP, which predicts per-pixel blur kernel to model defocus and camera motion blur. 
Though the inference stage does not involve the blur kernel estimation, it is no different from the training with regard to rendering time as it is based on volumetric rendering which takes several seconds to render a single image. 
DP-NeRF~\cite{lee2022deblurred} and PDRF~\cite{peng2022pdrf} further improved Deblur-NeRF, still they depend on volumetric rendering and are not free from the rendering cost.
Other approaches~\cite{dai2023hybrid,wang2022bad,sharpnerf} are bounded to addressing only one type of blur, either camera motion blur or defocus blur, and not aimed at solving the long rendering time. While these deblurring NeRFs successfully produce clean images from the blurry input images, there is room for improvement in terms of rendering time.
Thus, we propose a novel deblurring framework, Deblurring 3D Gaussian Splatting, which enables real-time sharp image rendering using a differentiable rasterizer and 3D Gaussians.

\section{Deblurring 3D Gaussian Splatting}
Based on the 3D-GS~\cite{kerbl20233d}, we generate 3D Gaussians, and each Gaussian is uniquely characterized by a set of the parameters, including 3D position $x$, opacity $\sigma$, and covariance matrix derived from quaternion $r$ scaling $s$. Every 3D Gaussian also contains spherical harmonics (SH) to represent view-dependent appearance. The input for the proposed method consists of camera poses and point clouds, which can be obtained through the structure from motion (SfM)~\cite{schonberger2016structure}, and a collection of images (possibly blurred). We employ an MLP that takes $x_j$, $r_j$, and $s_j$ which are 3D position, quaternion, and scaling of $j$-th Gaussian, respectively as inputs to deblur a scene. In case of modeling defocus blur, the MLP yields $\delta r_j$ and $\delta s_j$ which are the small scaling factors multiplied to $r$ and $s$, respectively. With new quaternion and scale, $r_j\cdot \delta r_j$ and $s_j \cdot \delta s_j$, the updated 3D Gaussians are subsequently fed to the tile-based rasterizer to rasterize the defocus blurred images. To address camera motion blur, the MLP outputs $\{(\delta x_j^{(i)}, \delta r_j^{(i)}, \delta s_j^{(i)})\}_{i=1}^{M}$ where $M$ is the number of the auxiliary sets of 3D Gaussians representing the moments of camera movement and $\delta x_j^{(i)}, \delta r_j^{(i)}, \delta s_j^{(i)}$ are $i$-th predicted position offset, scaling factor for scale, and scaling factor for quaternion of $j$-th 3D Gaussian, respectively. Rasterizer produces $M$ images from $M$ different sets of 3D Gaussians and we average them to obtain camera motion blurred image. The overview of our method is shown in Fig.~\ref{fig:workflow}.

\subsection{Differential Rendering via 3D Gaussian Splatting}
At the training time, the blurry images are rendered in a differentiable way and we use a gradient-based optimization to train our Deblurring 3D Gaussians. We adopt methods from \cite{kerbl20233d}, which proposes differentiable rasterization. Each 3D Gaussian is defined by its covariance matrix $\Sigma(r,s)$ with mean value in 3D world space $x$ as following: 
\begin{equation}
\label{formula:gaussian's formula}
    G(x, r, s)=e^{-\frac{1}{2} x^T\Sigma^{-1}(r,s) x}.
    \vspace{0.2cm}
\end{equation}
Besides $\Sigma(r,s)$ and $x$, 3D Gaussians are also defined with spherical harmonics coefficients (SH) to represent view-dependent appearance and opacity for alpha value. The covariance matrix is valid only when it satisfies positive semi-definite, which is challenging to constrain during the optimization. Thus, the covariance matrix is decomposed into two learnable components, a quaternion $r$ for representing rotation and $s$ for representing scaling, to circumvent the positive semi-definite constraint similar to the configuration of an ellipsoid. $r$ and $s$ are transformed into rotation matrix and scaling matrix, respectively, and construct $\Sigma(r,s)$ as follows: 
\begin{equation}
\label{formula:covariance decomposition}
    \Sigma(r,s) = R(r)S(s)S(s)^TR(r)^T,
    \vspace{0.2cm}
\end{equation}
where $R(r)$ is a rotation matrix given the rotation parameter $r$ and $S(s)$ is a scaling matrix from the scaling parameter $s$~\cite{kuipers1999quaternions}.
These 3D Gaussians are projected to 2D space~\cite{zwicker2002ewa} to render 2D images with following 2D covariance matrix $\Sigma'(r,s)$:
\begin{equation}
    \Sigma^{\prime}(r,s) = JW\Sigma(r,s) W^TJ^T,
    \vspace{0.2cm}
\end{equation}
where $J$ denotes the Jacobian of the affine approximation of the projective transformation, $W$ stands for the world-to-camera matrix. Each pixel value is computed by accumulating $N$ ordered projected 2D Gaussians overlaid on the each pixel with the formula:

\begin{equation}
\label{formula: splatting&volume rendering}
    C = \sum_{i\in N} T_i c_i \alpha_i   \hspace{0.5em} \text{  with  }  \hspace{0.5em} T_i = \prod_{j=1}^{i-1} (1-\alpha_j),
    \vspace{0.2cm}
\end{equation}
$c_i$ is the color of each point, and $T_i$ is the transmittance. $\alpha_i \in [0,1]$ defined by $1-\exp^{-\sigma_i\delta_i}$ where $\sigma_i$ and $\delta_i$ are the density of the point and the interval along the ray respectively. For further details, please refer to the original paper~\cite{kerbl20233d}.

\subsection{Deblurring 3D Gaussians}
\label{sec:deblurring_3dgs}
\subsubsection{Motivation} It is discussed in \cref{Eq:blur_convolution} that the pixels in images get blurred due to defocusing and camera motion, and this phenomenon is usually modeled through a convolution operation. 
Correspondingly, an image captured by a camera is the result of the convolution of the actual image and the PSF. Through convolution, which is the weighted summation of neighboring pixels, some pixels can affect the central pixel heavily depending on the weight. In other words, in the blurry imaging process, a pixel affects the intensity of neighboring pixels. This theoretical base motivates us to build our Deblurring 3D Gaussians framework. 

When handling defocus blur, we assume that big-sized 3D Gaussians cause the blur, while smaller 3D Gaussians correspond to the sharp image. This is because those with greater dispersion are affected by more neighboring information as they are responsible for wider regions in image space, so they can represent the interference of the neighboring pixels. Whereas the fine details in the 3D scene can be better modeled through the smaller 3D Gaussians. In the case of camera motion blur, we implicitly model the camera movement during the camera exposure time. In detail, we generate multiple auxiliary sets of 3D Gaussians which represent the discrete moment of the movement, by shifting the positions of the existing set of 3D Gaussians, and simulate camera motion blur. More details are described in the supplementary material.

\subsubsection{Defocus blur modeling}\label{sec:defocus} Following the aforementioned motivation, we learn to deblur by transforming the geometry of the 3D Gaussians. The geometry of the 3D Gaussians is expressed through the covariance matrix, which can be decomposed into the rotation and scaling factors as mentioned in Eq.~\ref{formula:covariance decomposition}. Therefore, our target is to change the rotation and scaling factors of 3D Gaussians in such a way that we can model the blurring phenomenon. To do so, we have employed an MLP that takes the position $x_j$, rotation $r_j$, scale $s_j$ of $j$-th 3D Gaussian, and viewing direction $v$ as inputs, and outputs $(\delta r_j, \delta s_j)$, as given by: 
\begin{equation}
\label{eq:mlp_defocus}
    (\delta r_j, \delta s_j) = \mathcal{F_\theta}\Big(\gamma(x_j), r_j, s_j, \gamma(v) \Big),
    \vspace{0.2cm}
\end{equation} 
where $\mathcal{F}_\theta$ denotes the MLP, and $\gamma$ denotes the positional encoding which is defined as:
\begin{equation}
    \gamma(p) = \big(\sin(2^k\pi p), \cos(2^k \pi p)\big)_{k=0}^{L-1},
    \vspace{0.2cm}
\end{equation}
where $L$ is the number of the frequencies, and the positional encoding is applied to each element of the vector $p$~\cite{nerf}.

Each scaling factor $(\delta r_j, \delta s_j)$ is scaled by $\lambda_s$, shifted by $(1 - \lambda_s)$ for optimization stability. Then the minima of them are clipped to 1.0 and element-wisely multiplied to $r_j$ and $s_j$, respectively, to obtain the transformed attributes as following:
\begin{align}
\label{eq:hat_r}
\hat{r}_j=r_j\cdot \text{min}(1.0, ~\lambda_s \delta r_j + (1-\lambda_s)), \\
\hat{s}_j=s_j\cdot \text{min}(1.0, ~\lambda_s \delta s_j + (1-\lambda_s)),
\label{eq:hat_s}
\end{align}
where function $\text{min}(\cdot,\cdot)$ returns smaller value. With these transformed attributes, we can construct the transformed 3D Gaussians $G(x_j, \hat{r}_j, \hat{s}_j)$, which is optimized during training to model the scene blurriness.
As $\hat{s}_j$ is greater than or equal to $s$, each 3D Gaussian of $G(x_j, \hat{r}_j, \hat{s}_j)$ has greater statistical dispersion than the original 3D Gaussian $G(x_j, r_j, s_j)$.
With the expanded dispersion of 3D Gaussian, it can represent the interference of the neighboring information which is a root cause of defocus blur.
In addition, $G(x_j, \hat{r}_j, \hat{s}_j)$ can model the blurry scene more flexibly as per-Gaussian $\delta r$ and $\delta s$ are estimated.
Defocus blur is spatially varying, which implies different regions have different levels of blurriness.
The scaling factors for 3D Gaussians that are responsible for a region with harsh defocus blur where various neighboring information in wide range is involved in, become bigger to better model a high degree of blurriness.
Meanwhile, those for 3D Gaussians on the sharp area are closer to 1.0 so that they have smaller dispersion and do not represent the influence of the nearby information.
Therefore, we can model defocus blur and rasterize defocus blurred image with $G(x_j, \hat{r}_j, \hat{s}_j)$.

At the time of inference, we use $G(x_j, r_j, s_j)$ to render the sharp images. As mentioned earlier, we assume that multiplying two different scaling factors to transform the geometry of 3D Gaussians can work as blur kernel and convolution in \cref{Eq:blur_convolution}. Thus, $G(x_j, r_j, s_j)$ can produce the images with clean and fine details. It is worth noting that since any additional scaling factors are not used to render the images at testing time, $F_\theta$ is not activated, so all steps required for the inference of Deblurring 3D-GS are identical to 3D-GS, which in turn enables real-time sharp image rendering. 

\subsubsection {Camera motion blur modeling}
We model camera motion blur with additional sets of 3D Gaussians. We adjust the geometry of each 3D Gaussian to simulate blur at training time, akin to defocus blur modeling. However, unlike defocus blur, camera motion blur occurs due to the physical movement of a camera. Every moment when the light hits the camera sensor, camera movement during the exposure time makes light intensities from multiple sources intermixed. We model such a phenomenon by adding small offsets to the position of each 3D Gaussian, and produce additional sets of 3D Gaussians to implicitly represent camera shake, and average clean images from different moments to simulate camera motion blur. Specifically, we first slightly change \cref{eq:mlp_defocus} to compute additional output $\delta x$, the offset for the position $x_j$ of a Gaussian, from $F_\theta$ as:
\begin{equation}
\label{eq:mlp}
    \{(\delta x_j^{(i)}, \delta r_j^{(i)}, \delta s_j^{(i)})\}_{i=1}^{M} = \mathcal{F_\theta}\Big(\gamma(x_j), r_j, s_j, \gamma(v) \Big),
    \vspace{0.2cm}
\end{equation}
where $M$ is the number of additional sets of 3D Gaussian to model the moments during the camera movement, $\delta x_j^{(i)}$ is $i$-th predicted position offset of $j$-th Gaussian.
$\delta r_j^{(i)} ~\text{and}~ \delta s_j^{(i)}$ are scaled, shifted, and clipped in the same manner to \cref{eq:hat_r} and \cref{eq:hat_s} for each.
Then we construct extra $M$ sets of 3D Gaussians $\{ \{(\hat{x}_j^{(i)},\hat{r}_j^{(i)},\hat{s}_j^{(i)})\}_{i=1}^M\}_{j=1}^{N_G}$ by shifting the positions and changing the geometry of the existing set of 3D Gaussians, where $N_G$ is the number of the current 3D Gaussians, $\hat{x}_j^{(i)}$ stands for the shifted position with scaled $\delta x_j^{(i)}$ by $\lambda_p$. $\hat{x}_j^{(i)} = x_j + \lambda_p\delta x_j^{(i)}$, and $\hat{r}_j^{(i)} = r_j \cdot \delta r_j^{(i)} ~\text{and}~ \hat{s}_j^{(i)} = s_j \cdot \delta s_j^{(i)}$ are computed in a similar way to compute them for defocus blur deblurring. Each set corresponds to 3D Gaussians observed from different camera viewpoints and we rasterize $M$ clean images from $M$ different sets of 3D Gaussians and then average them to obtain a single camera motion blurred image $I_b$ at the training time as following:
\begin{equation}
    I_b = \frac{1}{M}\sum_{i=1}^{M} I_i, \quad I_i = \texttt{Rasterize}(\{G(\hat{x}_j^{(i)},\hat{r}_j^{(i)},\hat{s}_j^{(i)})\}_{j=1}^{N_G}),
\end{equation}
where $I_i$ is a clean image generated by the predicted deltas.
Deblurring camera motion blur also does not require any MLP forwarding and rendering multiple images at the inference time as $G(x_j,r_j,s_j)$ learns the latent clean image. Thus, we can still enjoy rendering clean images from camera motion blurred input images in real-time manners, just like defocus blur deblurring. 


\begin{algorithm}[t]
\caption{Add Extra Points}
\begin{algorithmic}[0]
\Require{$\mathcal{P}$: Point cloud computed from SfM}
\Require{$K$: Number of the neighboring points to find}
\Require{$N_p$: Number of additional points to generate}
\Require{$t_d$: Minimum required distance between new point and existing point}

\State $P_\text{add}$ \textleftarrow GenerateRandomPoints($\mathcal{P}, N_p$) \Comment{Uniformly sample $N_p$ points}

\For{each $p$ in $P_\text{add}$}
\State $\mathcal{P}_\text{knn}$ \textleftarrow FindNearestNeighbors($\mathcal{P}, p, K$) \Comment{Get $K$ nearest points of $p$ from $\mathcal{P}$}
\State $\mathcal{P}_\text{valid} \leftarrow$ CheckDistance($\mathcal{P}_\text{knn}$, $p$, $t_d$) \Comment{Discard irrelevant neighbors}

\If {$|\mathcal{P}_\text{valid}| > 0 $ }
\State $p_c \leftarrow$ LinearInterpolate($\mathcal{P}_\text{valid}$, $p$) \Comment{Linearly interpolate neighboring colors}
\State AddToPointCloud($\mathcal{P}, p, p_c$)
\EndIf
\EndFor
\end{algorithmic}
\label{alg:add_pts}
\end{algorithm}

\subsection{Compensation for Sparse Point Cloud}
\begin{figure}[]
\begin{center}
\includegraphics[width=0.9\linewidth]{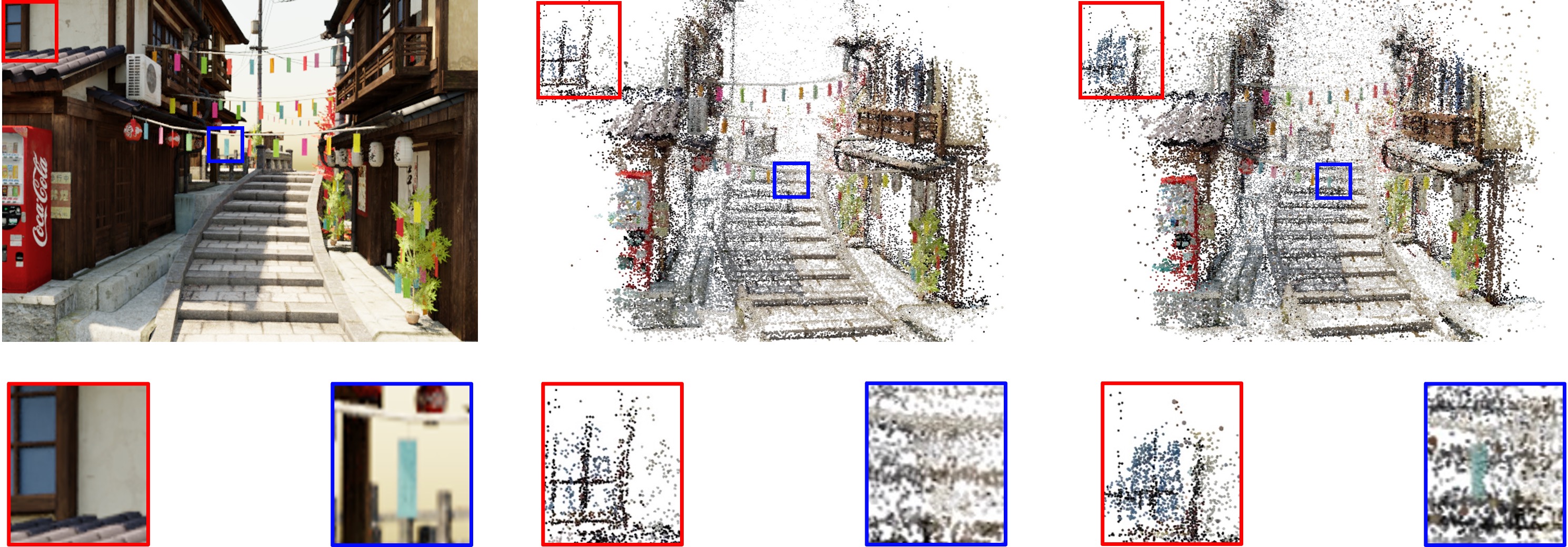}
\end{center}
\vspace{-0.5cm}
   \caption{Comparison on densifying point clouds during training. Left: Example training view. Middle: Point cloud at 5,000 training iterations without adding points. Right: Point cloud at 5,000 training iterations with adding extra points at 2,500 iterations.}
\label{fig:extra_points}
\end{figure}

3D-GS~\cite{kerbl20233d} constructs multiple 3D Gaussians from point clouds to model 3D scenes, and its reconstruction quality heavily relies on the initial point clouds. Point clouds are generally obtained from the structure-from-motion (SfM)~\cite{schoenberger2016sfm}, which extracts features from multi-view images and relates them to several 3D points. However, it can produce only sparse point clouds if the given images are blurry.
Even worse, if the scene has a large depth of field, which is prevalent in defocus blurry scenes, SfM hardly extracts any points that lie on the far end of the scene. To make a dense point cloud, we add extra points after $N_{st}$ iterations. $N_p$ points are sampled from a uniform distribution $U(\alpha, \beta)$ where $\alpha$ and $\beta$ are the minimum and maximum value of the position of the points from the existing point cloud, respectively. The color for each new point $p$ is assigned with the interpolated color $p_c$ from the nearest neighbors $\mathcal{P}_\text{knn}$ among the existing points using K-Nearest-Neigbhor (KNN)~\cite{knn}. We discard the points whose distance to the nearest neighbor exceeds the distance threshold $t_d$ to prevent unnecessary points from being allocated to the empty space. The process of adding supplementary points to the given point cloud is summarized in \cref{alg:add_pts}. \cref{fig:extra_points} shows that a point cloud with additional points has a dense distribution of points to represent the objects.

\begin{figure}[h]
\begin{center}
\includegraphics[width=4.8in, keepaspectratio]{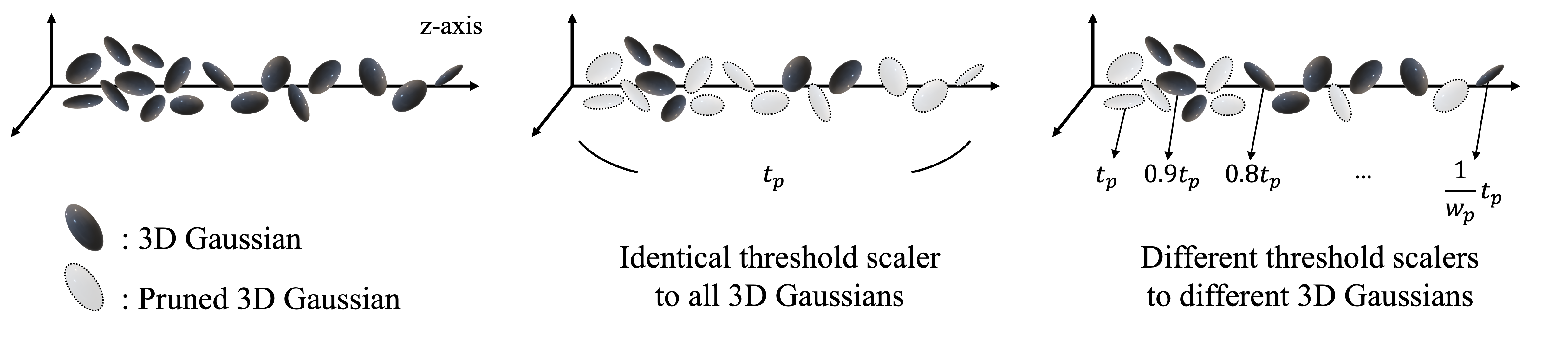}
\end{center}
   \caption{Comparison to pruning 3D Gaussians. Left: Given 3D Gaussians. Middle: Applying the pruning method proposed by 3D-GS which removes 3D Gaussians with the single threshold ($t_p$). Right: Our pruning method that discards unnecessary 3D Gaussians with different thresholds based on their depth.}
\label{fig:depth_pruning_concept}
\end{figure}

Furthermore, 3D-GS~\cite{kerbl20233d} effectively manages the number of 3D Gaussians through periodic adaptive density control, densifying and pruning 3D Gaussians. To compensate for the sparsity of 3D Gaussians lying on the far end of the scene, we prune 3D Gaussians depending on their positions. As the benchmark Deblur-NeRF dataset~\cite{ma2022deblur} consists of only forward-facing scenes, the z-axis value of each point can be a relative depth from any viewpoint. As shown in \cref{fig:depth_pruning_concept}, we prune out less 3D Gaussians placed on the far edge of the scene to preserve more points located at the far plane, relying on the relative depth. Specifically, the pruning threshold $t_p$ is scaled by $\frac{1}{w_p}$ where $w_p$ is determined depending on the relative depth, and the lowest threshold is applied to the farthest point. 

\section{Experiments}
We compared our method against the state-of-the-art deblurring approaches in neural rendering: Deblur-NeRF~\cite{ma2022deblur}, Sharp-NeRF~\cite{sharpnerf}, DP-NeRF~\cite{Lee_2023_CVPR}, PDRF~\cite{peng2022pdrf} and original 3D Gaussians Splatting (3D-GS)~\cite{kerbl20233d} and image-based deblurring which deblurs training images first using Restormer~\cite{restormer} and then trains 3D-GS with them. We evaluated the performance on the benchmark Deblur-NeRF dataset~\cite{ma2022deblur} that includes both synthetic and real images captured with either camera motion blur or defocus blur.
\subsection{Experimental Settings}
We use Adam optimizer~\cite{KingBa15} and set the learning rate for MLP to $1e-3$, that for the position of 3D Gaussians to $1.6e-3$. 3D Gaussian pruning threshold ($t_p$) and densification threshold are $5e-3$ and $2e-4$ respectively, for the real defocus blur dataset and $1e-2$ and $5e-4$ for the real camera motion blur dataset. The rest hyperparameters are identical to those of 3D-GS. We use an MLP with a depth of 4 layers. The first 3 layers are shared for all deltas and features from the shared layers are fed to each of 3 single layer (i.e., 1 layer head for each delta) that produces $\delta x, \delta r$, and $\delta s$ respectively. All layers have 64 hidden units, adopt ReLU activation for non-linearity, and are initialized with Xavier initialization~\cite {glorot2010understanding}. Both $\lambda_p$ and $\lambda_s$ are set to $1e-2$.
For adding extra points to compensate for the sparse point cloud, we set the addition start iteration $N_{st}$ to 2,500, the number of supplementing points $N_p$ is proportional to the extent of the point cloud, at most 200,000, further explained in the supplementary material. The number of neighbors $K$ is 4, and the minimum distance threshold $t_d$ is 2. In terms of depth-based pruning,  the pruning threshold multiplier $w_p$ is set to 3. We set $M$ for camera motion deblurring to 5 and the total iteration for training is 20,000.
All the experiments were conducted on NVIDIA RTX 4090 GPU.
\label{sec:experiments}


\subsection{Results and Comparisons}
In this section, we provide the outcomes of our experiments, presenting a thorough analysis of both qualitative and quantitative results. Our evaluation framework encompasses a diverse set of metrics to show a comprehensive assessment of the experimental results. Primarily, we rely on established metrics such as the Peak Signal-to-Noise Ratio (PSNR), Structural Similarity Index (SSIM), and Frames Per Second (FPS).

As shown in ~\cref{tab:res_imgref} and ~\cref{fig:curve} our method is on par with the state-of-the-art model in PSNR and achieve state-of-the-art performance evaluated under SSIM on the real defocus blur dataset. \cref{tab:real_motion_psnr} further shows that the proposed method attains state-of-the-art performance on real camera motion blur dataset, under all metrics. At the same time, the proposed method can still enjoy real-time rendering, with a noticeable FPS, while other deblurring models cannot. 
\cref{fig:qualitative_real_motion} shows the qualitative results on real camera motion blur dataset. We can see that ours can produce sharp and fine details, though 3D-GS fails to reconstruct those details. The qualitative and quantitative results on the rest datasets, more experiments and ablation studies are delivered in supplementary materials.

\begin{table*}[h!]
    \centering
    \vspace{-0.2cm}
    \caption{Quantitative results on real defocus blur dataset. We color each cell as \colorbox{orange!50}{best} and \colorbox{yellow!50}{second best}.}
    \resizebox{\linewidth}{!}{
    \begin{tabular}{c|cc|cc|cc|cc|cc|cc|c}
    \hline
         & \multicolumn{2}{c|}{Cake}& \multicolumn{2}{c|}{Caps} &  \multicolumn{2}{c|}{Cisco} & \multicolumn{2}{c|}{Coral} & \multicolumn{2}{c|}{Cupcake}& \\
     & PSNR$\uparrow$ & SSIM$\uparrow$ & PSNR$\uparrow$ & SSIM$\uparrow$ & PSNR$\uparrow$ & SSIM$\uparrow$ & PSNR$\uparrow$ & SSIM$\uparrow$ & PSNR$\uparrow$ & SSIM$\uparrow$ \\
     \hline
     \hline
    NeRF~\cite{nerf} & 24.42 & 0.7210 & 22.73 & 0.6312 & 20.72 & 0.7217 & 19.81 & 0.5658 & 21.88 & 0.6809 \\
    3D-GS~\cite{kerbl20233d} & 20.16 & 0.5903 & 19.08 & 0.4355 & 20.01 & 0.6931 & 19.50 & 0.5519 & 21.53 & 0.6794  \\
    Restormer~\cite{restormer} + 3D-GS & 21.48 & 0.6271 & 21.80 & 0.5956 & 19.95 & 0.6778  & \cellcolor{orange!50}20.63 & \cellcolor{orange!50} 0.6125 & \cellcolor{orange!50}23.18 & \cellcolor{orange!50} 0.7523  \\
    \hline
    Deblur-NeRF~\cite{ma2022deblur} & 26.27 & 0.7800 & 23.87 & \cellcolor{yellow!50}0.7128 & 20.83 & \cellcolor{yellow!50}0.7270 & 19.85 & 0.5999 & 22.26 & 0.7219 \\
    Sharp-NeRF~\cite{sharpnerf} & 26.23 & 0.7799 & 23.98 & 0.7098 & \cellcolor{orange!50}20.88 & 0.7269 & 20.07 & 0.5999 & 22.75 & 0.7376 \\
    DP-NeRF~\cite{lee2022deblurred} & 26.16 & 0.7781 & 23.95 & 0.7122 & 20.73 & 0.7260 & \cellcolor{yellow!50}20.11 & \cellcolor{yellow!50}0.6107 & 22.80 & 0.7409 \\
    PDRF-10~\cite{peng2022pdrf} & \cellcolor{orange!50}27.06 & \cellcolor{orange!50}0.8032 & \cellcolor{yellow!50}24.06 & 0.7102 & 20.68 & 0.7239 & 19.61 & 0.5894 & \cellcolor{yellow!50}22.95 & \cellcolor{yellow!50}0.7421 \\
    Ours & \cellcolor{yellow!50}26.88 & \cellcolor{yellow!50}0.8026 & \cellcolor{orange!50}24.50 & \cellcolor{orange!50}0.7428 & \cellcolor{yellow!50}20.83 & \cellcolor{orange!50}0.7321 & 19.78 & 0.6080 & 22.11 & 0.7344 \\
    \hline
         & \multicolumn{2}{c|}{Cups}& \multicolumn{2}{c|}{Daisy} &  \multicolumn{2}{c|}{Sausage} & \multicolumn{2}{c|}{Seal} & \multicolumn{2}{c|}{Tools}& \multicolumn{2}{c|}{Average} & FPS \\
     & PSNR$\uparrow$ & SSIM$\uparrow$ & PSNR$\uparrow$ & SSIM$\uparrow$ & PSNR$\uparrow$ & SSIM$\uparrow$ & PSNR$\uparrow$ & SSIM$\uparrow$ & PSNR$\uparrow$ & SSIM$\uparrow$ & PSNR$\uparrow$ & SSIM$\uparrow$ & $\uparrow$\\
     \hline
     \hline
    NeRF~\cite{nerf} & 25.02 & 0.7581 & 22.74 & 0.6203 & 17.79 & 0.4830 & 22.79 & 0.6267 & 26.08 & 0.8523 & 22.40 & 0.6661 & $<$ 1 \\
    3D-GS~\cite{kerbl20233d} & 20.55 & 0.6459 & 20.96 & 0.6004 & 17.83 & 0.4718 & 22.25 & 0.5905 & 23.82 & 0.805 & 20.57 & 0.6064 & \cellcolor{yellow!50}788 \\
    Restormer~\cite{restormer} + 3D-GS & 21.89 & 0.7160 & 21.17 & 0.6196 & 17.99 & 0.5060 & 25.05 & 0.7364 & 24.63 & 0.8235 & 21.78 & 0.6669  & 775 \\
    \hline
    Deblur-NeRF~\cite{ma2022deblur} & 26.21 & 0.7987 & 23.52 & 0.6870 & 18.01 & 0.4998 & 26.04 & 0.7773 & 27.81 & 0.8949 & 23.46 & 0.7199 & $<$ 1 \\
    Sharp-NeRF~\cite{sharpnerf} & 25.34 & 0.7783 & 23.66 & 0.7094 & 18.77 & 0.5482 & 25.82 & 0.7769 & 27.98 & \cellcolor{yellow!50}0.9032 & 23.55 & 0.7242 & $<$ 1 \\
    DP-NeRF~\cite{lee2022deblurred} & \cellcolor{orange!50}26.75 & \cellcolor{yellow!50}0.8136 & \cellcolor{yellow!50}23.79 & 0.6971 & 18.35 & 0.5443 & 25.95 & 0.7779 & \cellcolor{orange!50}28.07 & 0.8980 & 23.67 & 0.7299 & $<$ 1 \\
    PDRF-10~\cite{peng2022pdrf} & \cellcolor{yellow!50}26.39 & 0.8066 & \cellcolor{orange!50}24.49 & \cellcolor{orange!50}0.7426 & \cellcolor{yellow!50}18.94 & \cellcolor{yellow!50}0.5686 & \cellcolor{orange!50}26.36 & \cellcolor{yellow!50}0.7959 & \cellcolor{yellow!50}28.00 & 0.8995 & \cellcolor{orange!50}23.85 & \cellcolor{yellow!50}0.7382 & $<$ 1 \\
    Ours & 26.28 & \cellcolor{orange!50}0.8235 & 23.54 & \cellcolor{yellow!50}0.7310 & \cellcolor{orange!50}18.99 & \cellcolor{orange!50}0.5705 & \cellcolor{yellow!50}26.18 & \cellcolor{orange!50}0.8166 & 27.96 & \cellcolor{orange!50}0.9098 & \cellcolor{yellow!50}23.71 & \cellcolor{orange!50}0.7471 & \cellcolor{orange!50}804 \\
    \hline
    \end{tabular}
    }
    \label{tab:res_imgref}
\end{table*}

\begin{table*}[h!]
    \centering
    \vspace{-0.2cm}
    \caption{Quantitative results on real camera motion blur dataset tested under PSNR, SSIM and FPS. We color each cell as \colorbox{orange!50}{best} and \colorbox{yellow!50}{second best}. 
    }
    \resizebox{\linewidth}{!}{
    \begin{tabular}{c|cc|cc|cc|cc|cc|cc|c}
    \hline
         & \multicolumn{2}{c|}{Ball}& \multicolumn{2}{c|}{Basket} &  \multicolumn{2}{c|}{Buick} & \multicolumn{2}{c|}{Coffee} & \multicolumn{2}{c|}{Decoration}& \\
     & PSNR$\uparrow$ & SSIM$\uparrow$ & PSNR$\uparrow$ & SSIM$\uparrow$ & PSNR$\uparrow$ & SSIM$\uparrow$ & PSNR$\uparrow$ & SSIM$\uparrow$ & PSNR$\uparrow$ & SSIM$\uparrow$ \\
     \hline
     \hline
    NeRF~\cite{nerf} & 24.08 & 0.6237 & 23.72 & 0.7086 & 21.59 & 0.6325 & 26.47 & 0.8064 & 22.39 & 0.6609 \\
    3D-GS~\cite{kerbl20233d} & 22.99 & 0.6206 & 23.11 & 0.6833 & 21.22 & 0.6519 & 23.53 & 0.6995 & 20.45 & 0.6239  \\
    Restormer~\cite{restormer} + 3D-GS & 23.85 & 0.6498 & 23.75 & 0.7208 & 21.42 & 0.6949 & 23.94 & 0.7235 & 20.98 & 0.6840 \\  
    \hline
    Deblur-NeRF~\cite{ma2022deblur} & 27.36 & \cellcolor{yellow!50}{0.7656} & 27.67 & 0.8449 & 24.77 & 0.7700 & 30.93 & 0.8981 & 24.19 & 0.7707 \\
    DP-NeRF~\cite{lee2022deblurred} & 27.20 & 0.7652 & 27.74 & 0.8455 & \cellcolor{yellow!50}{25.70} & \cellcolor{yellow!50}{0.7922} & 31.19 & \cellcolor{yellow!50}{0.9049} & \cellcolor{yellow!50}{24.31} & \cellcolor{yellow!50}{0.7811}\\
    PDRF-10~\cite{peng2022pdrf} & \cellcolor{yellow!50}{27.96} & 0.7365 & \cellcolor{orange!50}{28.82} & \cellcolor{yellow!50}{0.8465} & 25.52 & 0.7742 & \cellcolor{yellow!50}{31.55} & 0.8627 & 23.26 & 0.7164 \\
    Ours & \cellcolor{orange!50}{28.27} & \cellcolor{orange!50}{0.8233} & \cellcolor{yellow!50}{28.42} & \cellcolor{orange!50}{0.8713} & \cellcolor{orange!50}{25.95} & \cellcolor{orange!50}{0.8367} & \cellcolor{orange!50}{32.84} & \cellcolor{orange!50}{0.9312} & \cellcolor{orange!50}{25.87} & \cellcolor{orange!50}{0.8540} \\
    \hline
         & \multicolumn{2}{c|}{Girl}& \multicolumn{2}{c|}{Heron} &  \multicolumn{2}{c|}{Parterre} & \multicolumn{2}{c|}{Puppet} & \multicolumn{2}{c|}{Stair}& \multicolumn{2}{c|}{Average} & FPS \\
     & PSNR$\uparrow$ & SSIM$\uparrow$ & PSNR$\uparrow$ & SSIM$\uparrow$ & PSNR$\uparrow$ & SSIM$\uparrow$ & PSNR$\uparrow$ & SSIM$\uparrow$ & PSNR$\uparrow$ & SSIM$\uparrow$ & PSNR$\uparrow$ & SSIM$\uparrow$ & $\uparrow$\\
     \hline
     \hline
    NeRF~\cite{nerf} & 20.07 & 0.7075 & 20.50 & 0.5217 & 23.14 & 0.6201 & 22.09 & 0.6093 & 22.87 & 0.4561 & 22.69 & 0.6347 & $<$ 1 \\
    3D-GS~\cite{kerbl20233d} & 19.72 & 0.7031 & 19.26 & 0.4767 & 22.22 & 0.5813 & 22.18 & 0.6362 & 21.88 & 0.4789 & 21.66 & 0.6154 & \cellcolor{yellow!50}{734} \\
    Restormer~\cite{restormer} + 3D-GS & 19.71 & 0.7151 & 19.68 & 0.5615 & 22.60 & 0.6364 & 22.19 & 0.6608 & 22.66 & 0.5735 & 22.08 & 0.6620 & 708 \\
    \hline
    Deblur-NeRF~\cite{ma2022deblur} & 22.27 & 0.7976 & 22.63 & 0.6874 & 25.82 & 0.7597 & 25.24 & 0.7510 & 25.39 & 0.6296 & 25.63 & 0.7675 & $<$ 1 \\
    DP-NeRF~\cite{lee2022deblurred} & \cellcolor{yellow!50}{23.33} & \cellcolor{yellow!50}{0.8139} & 22.88 & \cellcolor{yellow!50}{0.6930}& \cellcolor{yellow!50}{25.86} & \cellcolor{yellow!50}{0.7665} & \cellcolor{yellow!50}{25.25} & \cellcolor{yellow!50}{0.7536} & 25.59 & \cellcolor{yellow!50}{0.6349} & 25.91 & \cellcolor{yellow!50}{0.7751} & $<$ 1 \\
    PDRF-10~\cite{peng2022pdrf} & \cellcolor{orange!50}{23.78} & 0.8120 & \cellcolor{yellow!50}{22.90} & 0.6590 & 25.19 & 0.7233 & 25.06 & 0.7326 & \cellcolor{yellow!50}{25.73} & 0.5722 & \cellcolor{yellow!50}{25.98} & 0.7245 & $<$ 1 \\
    Ours & 23.26 & \cellcolor{orange!50}{0.8390} & \cellcolor{orange!50}{23.14} & \cellcolor{orange!50}{0.7438} & \cellcolor{orange!50}{26.17} & \cellcolor{orange!50}{0.8144} & \cellcolor{orange!50}{25.67} & \cellcolor{orange!50}{0.8051} & \cellcolor{orange!50}{26.46} & \cellcolor{orange!50}{0.7050} & \cellcolor{orange!50}{26.61} & \cellcolor{orange!50}{0.8224} & \cellcolor{orange!50}{961} \\
    \hline
    \end{tabular}
    }
    \vspace{-0.5cm}
    \label{tab:real_motion_psnr}
\end{table*}

\begin{figure*}[p]
    \centering
    \includegraphics[width=\textwidth, keepaspectratio]{figs/qualitative_real_motion.jpg}
    \caption{Qualitative results on real camera motion blur dataset.}
    \label{fig:qualitative_real_motion}
    
\end{figure*}



 \section{Limitations \& Future Works}
NeRF-based deblurring methods~\cite{ma2022deblur,Lee_2023_CVPR,peng2022pdrf}, which are developed under the assumption of volumetric rendering, are not easily applicable to rasterization-based 3D-GS~\cite{kerbl20233d}. However, they can be compatible to rasterization by optimizing their MLP to deform kernels in the space of the rasterized image instead of letting MLP deform the rays and kernels in the world space. Although it is an interesting direction, it will incur additional costs for interpolating pixels and just implicitly transform the geometry of 3D Gaussians. Therefore, we believe that it will not be an optimal way to model scene blurriness using 3D-GS~\cite{kerbl20233d}. 

\section{Conclusion}
We proposed Deblurring 3D-GS, the first deblurring algorithm for 3D-GS. We adopted a small MLP that transforms the 3D Gaussians to model the scene blurriness. We also further facilitated deblurring by complementing more points on sparse point clouds. We validated that our method can deblur the scene while still enjoying the real-time rendering with FPS $>$ 800. This is because we use the MLP only during the training time, and the MLP is not involved in the inference stage, keeping the inference stage identical to the 3D-GS.

\section*{Acknowledgments}
This research was supported in parts by the grant (RS-2023-00245342) from the Ministry of Science and ICT of Korea through the National Research Foundation (NRF) of Korea, Institute of Information and Communication Technology Planning Evaluation (IITP) grants for the AI Graduate School program (IITP-2019-0-00421), and the Culture, Sports, and Tourism R\&D Program through the Korea Creative Content Agency grant funded by the Ministry of Culture, Sports and Tourism in 2024 (Project Name: Research on neural watermark technology for copyright protection of generative AI 3D content, RS-2024-00348469). This work was also supported by Samsung Research Funding \& Incubation Center of Samsung Electronics under Project Number SRFC-IT2401-01.

%
%
\bibliographystyle{splncs04}
\bibliography{main}


\clearpage
\appendix
\renewcommand\thesection{\Alph{section}}
\setcounter{section}{0}
\begin{center}
\Large{\bf{Deblurring 3D Gaussian Splatting: \\ Supplementary Materials}}\par\vspace{3ex}
\end{center}

\section{Additional Method Details}
\subsubsection{Defocus blur modeling}
According to the thin lens law~\cite{hecht2012optics}, the scene points that lie at the focal distance of the camera make a sharp image at the imaging plane. On the other hand, any scene points that are not at the focal distance will make a blob instead of a sharp point on the imaging plane, and it produces a defocus blurred image. 
If the separation from the focal distance of a scene point is large, it produces a blob of a large area, which corresponds to severe defocus blur.
We assume that 3D Gaussians with greater dispersion (greater scale $s$) can represent the scene points not being located at the focal distance, while those with smaller scale $s$ can model the points placed at the focal distance. 
Table~\ref{tab:transformation_of_r_s} shows that during training, as compared to testing, 3D Gaussians of larger values of scale have been used to rasterize the scene. This indicates that larger values of scale are needed to adjust the 3D Gaussians to rasterize the blurred (training) images. While, in contrast, smaller values of the attributes demonstrate that the smaller-sized Gaussians are more suitable to model the fine details that are present in the sharp (testing) images.

\begin{table}[h!]
    \centering
    \caption{Scale transformation of 3D Gaussians to model the defocus blur measured on real defocus blur dataset.}
    \begin{tabular}{c|c|c}
    \hline
      & Train ($\hat{s}_j$) & Test ($s_j$) \\
    \hline\hline
    Average & 0.3232 & 0.3079 \\
    \hline
    \end{tabular}
    \label{tab:transformation_of_r_s}
\end{table}

\subsubsection{Camera motion blur modeling}
\label{sec:motivation}
\begin{figure}[t]
\begin{center}
\includegraphics[width=0.9\linewidth]{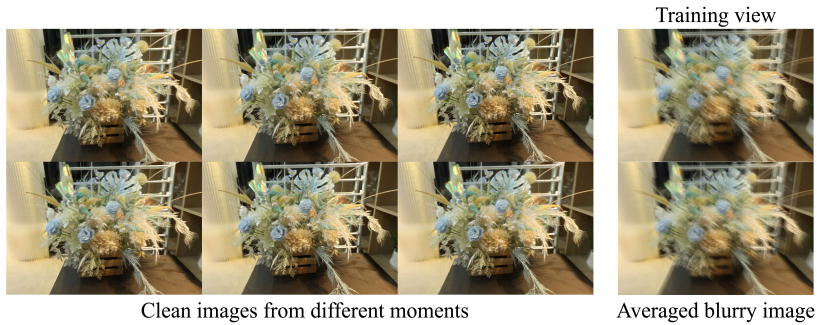}
\end{center}
\vspace{-0.5cm}
   \caption{Camera motion blur modeling during training. Clean images at different moments of the camera movement during exposure time are rasterized and averaged to a single image to model the camera motion blur.}
\label{fig:neighbors}
\end{figure}

Camera motion blur occurs due to the camera shake during the exposure time of a photograph. When a camera moves while capturing an image, the rays from different points within the scene strike different areas of the camera sensor at different moments and are intermixed, so the blurry final image is obtained~\cite{fergus2006removing}. 
Consequently, the blurry final image is acquired as each point in the subject is captured slightly different region on the sensor. Meanwhile, images at all the instants when the rays hit the camera sensor are sharp, without any inherent blur. Therefore, we train the 3D Gaussians to model clean representations at various moments the lights reach the sensor, rasterize $M$ clean images from $M$ different moments, and then average them to simulate the camera movement during training time. \cref{fig:neighbors} shows the clean images rasterized at different moments and averaged blurry image when $M$ is 6.

\subsubsection{Selective defocus blurring}
\begin{figure}[]
\begin{center}
\includegraphics[width=3.5in, keepaspectratio]{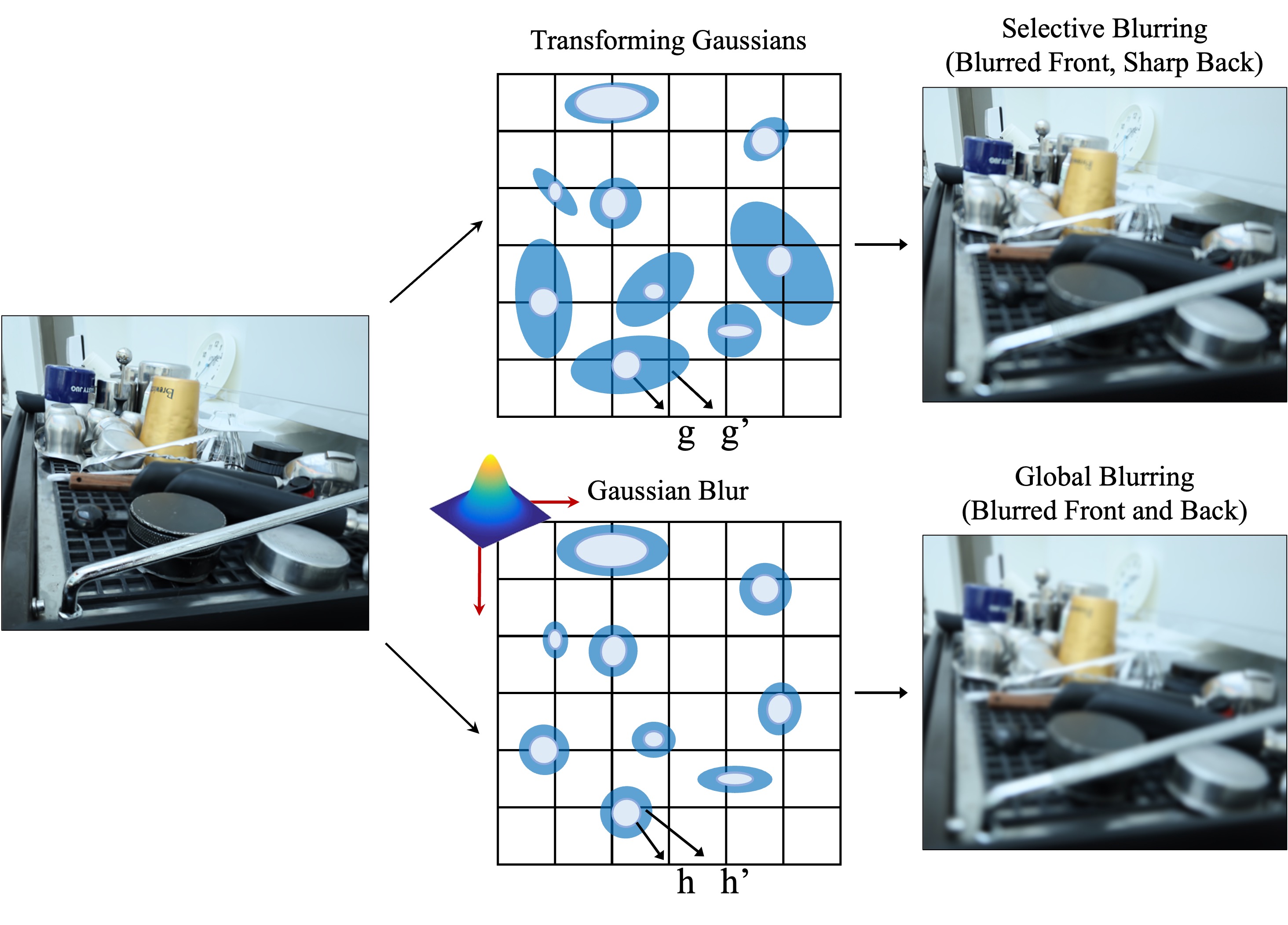}
\end{center}
\vspace{-0.5cm}
   \caption{Comparison to normal Gaussian blur kernel. Top row: It shows the proposed method. $g$ is the Gaussian before the transformation, and $g'$ is the Gaussian after the transformation. Since the different transformations can be applied to different Gaussian, ours can selectively blur the images depending on the scene; it can only blur the front parts of the scene. Bottom row: It describes a normal Gaussian blur kernel where $h$ is the Gaussian, and $h'$ is the Gaussian after applying a normal Gaussian blur kernel. Simply applying a normal Gaussian blur kernel is not capable of handling different parts of the image differently, thereby uniformly blurring the entire image.}
\label{fig:deltas}
\end{figure}

\begin{figure}[]
\begin{center}
\includegraphics[width=\textwidth, keepaspectratio]{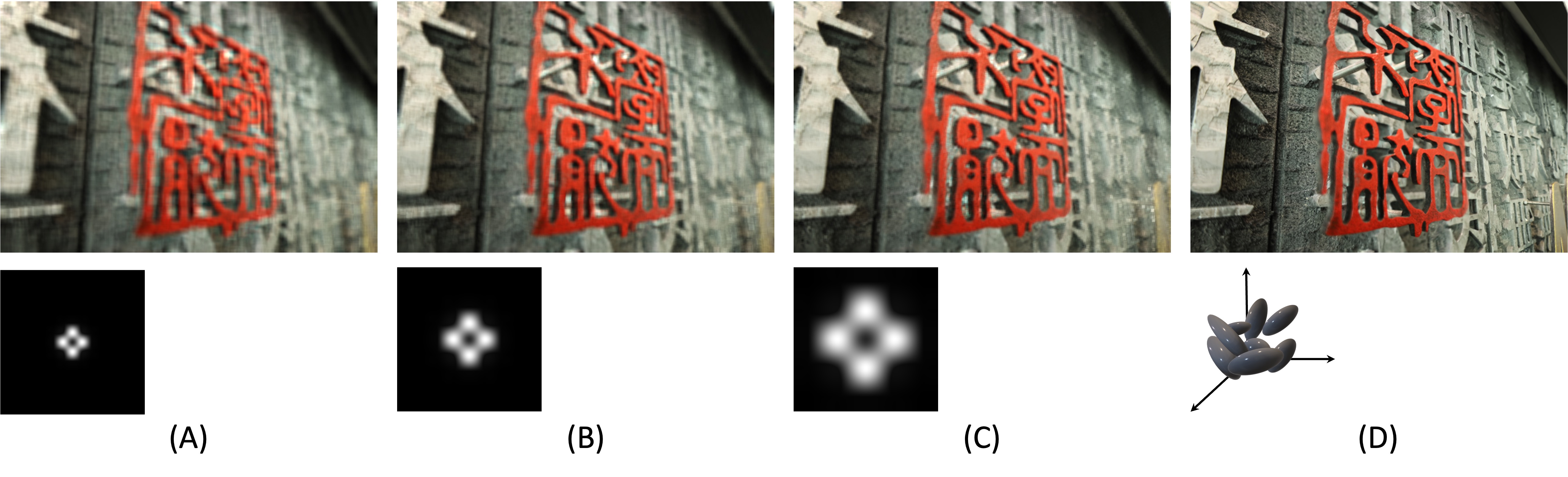}
\end{center}
\vspace{-0.5cm}
   \caption{Defocus-deblurred images with different sizes of normal Gaussian blur kernels and the proposed method. (A), (B), (C): 15$\times$15, 9$\times$9, and 5$\times$5 Gaussian blur kernel are in use to deblur, respectively, and the bottom row shows the visualization of each kernel whose values are inverted for better visibility (D): Proposed method which transforms geometry of each 3D Gaussian.
   }
\label{fig:gaussian_blur_kernel}
\end{figure}

\begin{figure}[]
\begin{center}
\includegraphics[width=0.9\linewidth]{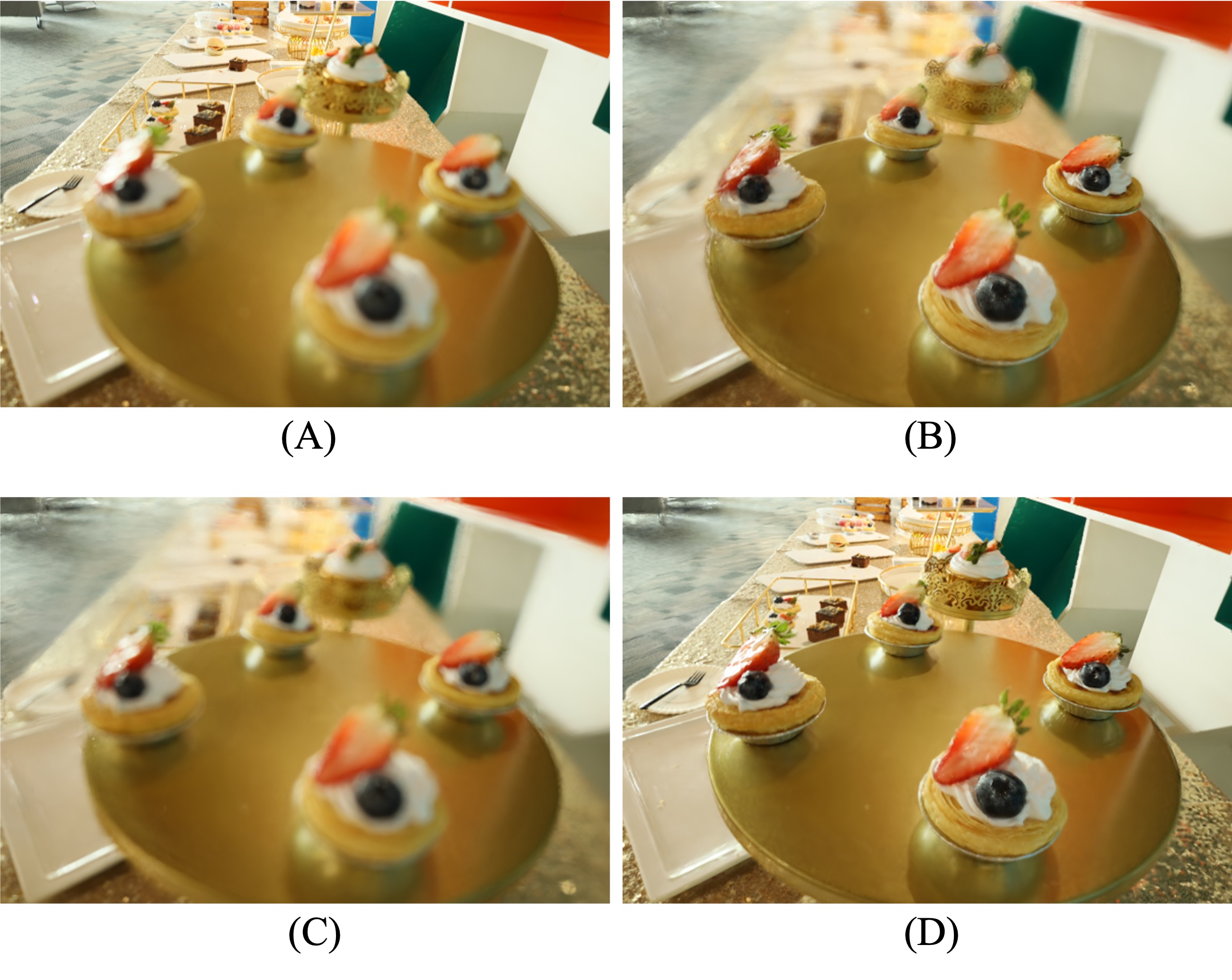}
\end{center}
\vspace{-0.5cm}
   \caption{Selective Gaussian blur adjustment. As delineated in ~\cref{fig:deltas}, our methodology adeptly harnesses the $\delta r_j, \delta s_j$ both emanating from compact Multi-Layer Perceptrons (MLP), enabling the inversion of Gaussian blur regions or the comprehensive modulation of overall blurriness and sharpness. With the Transformation of $\delta r_j, \delta s_j$, our framework facilitates the nuanced blurring of proximal regions akin to A, as well as the deft blurring of distant locales akin to B. Furthermore, it offers the capability to manipulate the global blurriness or sharpness, exemplified by adjustments akin to C and D.}
\label{fig:selective}
\end{figure}

The proposed method can handle the training images arbitrarily blurred in various parts of the scene. Since we predict $(\delta s_j)$ for each Gaussian, we can selectively enlarge the covariances of Gaussians where the parts in the training images are blurred. Such transformed per-Gaussian covariance is projected to 2D space and it can act as pixel-wise blur kernel at the image space. It is noteworthy that applying different shapes of blur kernels to different pixels plays a pivotal role in modeling scene blurriness since blurriness spatially varies. This flexibility enables us to effectively implement deblurring capability in 3D-GS~\cite{kerbl20233d}. On the other hand, a naive approach to blurring the rendered image is simply to apply a Gaussian kernel. As shown in Fig.~\ref{fig:deltas}, this approach will blur the entire image, not blur pixel-wisely, resulting in blurring parts that should not be blurred for training the model. Even if a learnable Gaussian kernel is applied, optimizing the mean and variance of the Gaussian kernel, a single type of blur kernel is limited in its expressivity to model the complicatedly blurred scene and is optimized to model the average blurriness of the scene from averaging loss function which fails to model blurriness morphing in each pixel. Not surprisingly, the Gaussian blur is a special case of the proposed method. If we predict one $(\delta s_j)$ for all 3D Gaussians, then it will similarly blur the whole image. \cref{fig:gaussian_blur_kernel} shows that the proposed method successfully deblur the defocus blur, while normal Gaussian blur kernel approaches fail. Moreover, transforming per-Gaussian covariance allows to adjust scene blurriness arbitrarily as shown in \cref{fig:selective}.

\subsubsection{Visualization}
\begin{figure}[]
\begin{center}
\includegraphics[width=\textwidth, keepaspectratio]{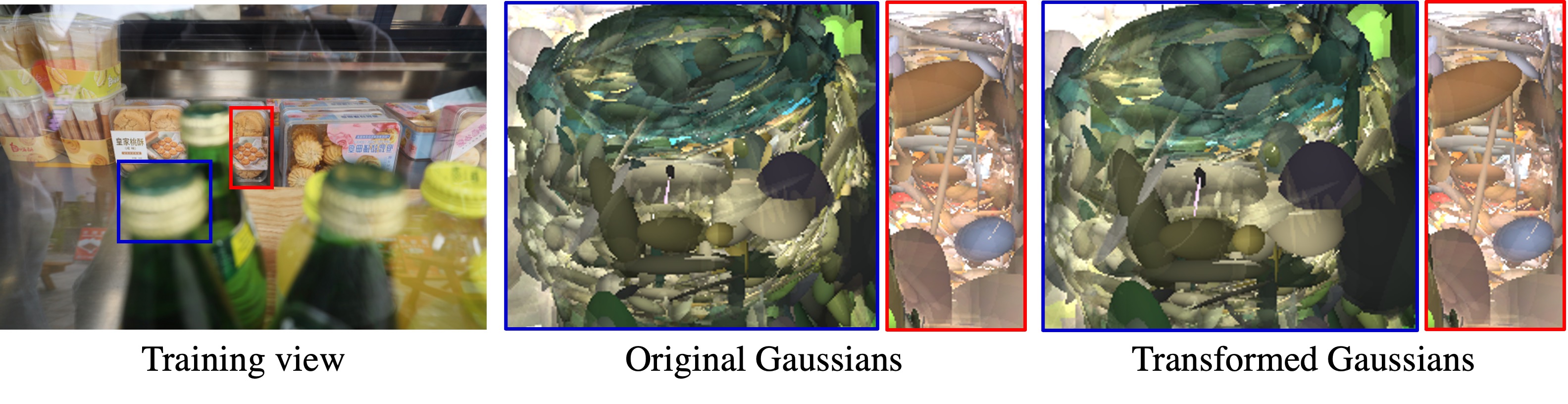}
\end{center}
\vspace{-0.5cm}
   \caption{Gaussians visualization for defocus blur.}
\label{fig:gaussian_visualization}
\end{figure}

\begin{figure}[]
\begin{center}
\includegraphics[width=\textwidth, keepaspectratio]{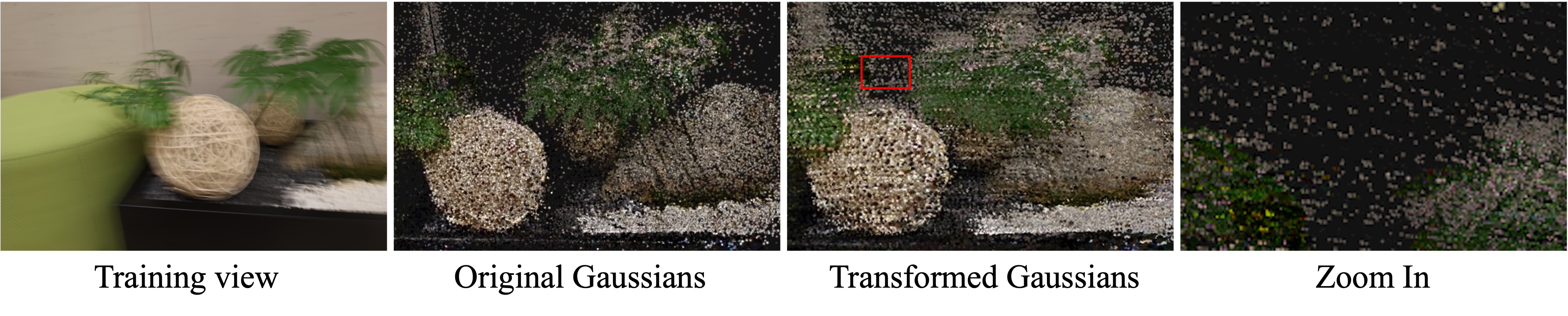}
\end{center}
\vspace{-0.5cm}
   \caption{Point cloud visualization for camera motion blur.}
\label{fig:point_cloud_visualization}
\end{figure}

\cref{fig:gaussian_visualization} visualizes the original and transformed 3D Gaussians for defocus blur. With a given view whose near plane is defocused, the transformed 3D Gaussians show larger scales than those of the original 3D Gaussians to model defocus blur on the near plane (blur-bordered images). Meanwhile, the transformed 3D Gaussians keep very similar shapes to the original ones for sharp objects in the far plane (red-bordered images). \cref{fig:point_cloud_visualization} depicts point clouds of the original 3D Gaussians and transformed 3D Gaussians. The point cloud of the transformed 3D Gaussians exhibits the camera movements when it moves left to right.

\subsubsection{Adding extra points}
The number of supplementing points $N_p$ is determined based on the extent of the point cloud as follows:
\begin{equation}
    N_p = \text{min}(\frac{\text{prod}(Q_{max} - Q_{min})}{c^3}, N_{max}),
\end{equation}
where $\text{prod}(\cdot)$ returns a product of all values in a vector, $Q_{max} \in \mathbb{R}^3$ and $Q_{min} \in \mathbb{R}^3$ are the maximum and minimum values of the positions of the points in the point cloud along x-, y-, and z-axis, respectively, and $c$ is a constant for stability which is set to 1.1. $N_{max}$ is 200,000 which restricts the number of points to be added.

\section{Additional Experiments}
\subsubsection{More results}
\begin{figure*}[p]
    \centering
    \includegraphics[width=\textwidth, keepaspectratio]{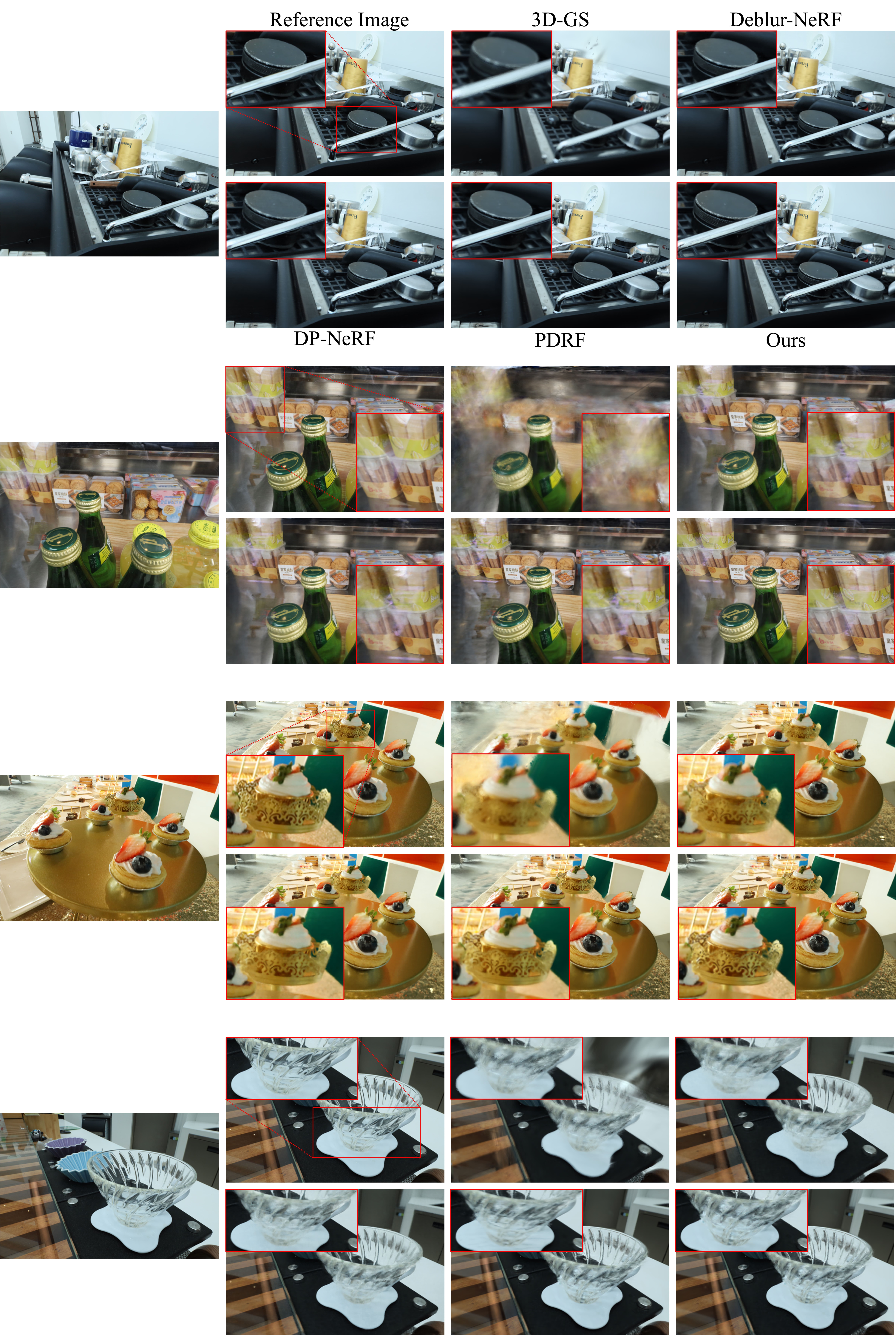}
    \caption{Qualitative results on real defocus blur dataset.}
    \label{fig:qualitative_real}
\end{figure*}

\begin{figure*}[p]
    \centering
    \includegraphics[width=\textwidth, keepaspectratio]{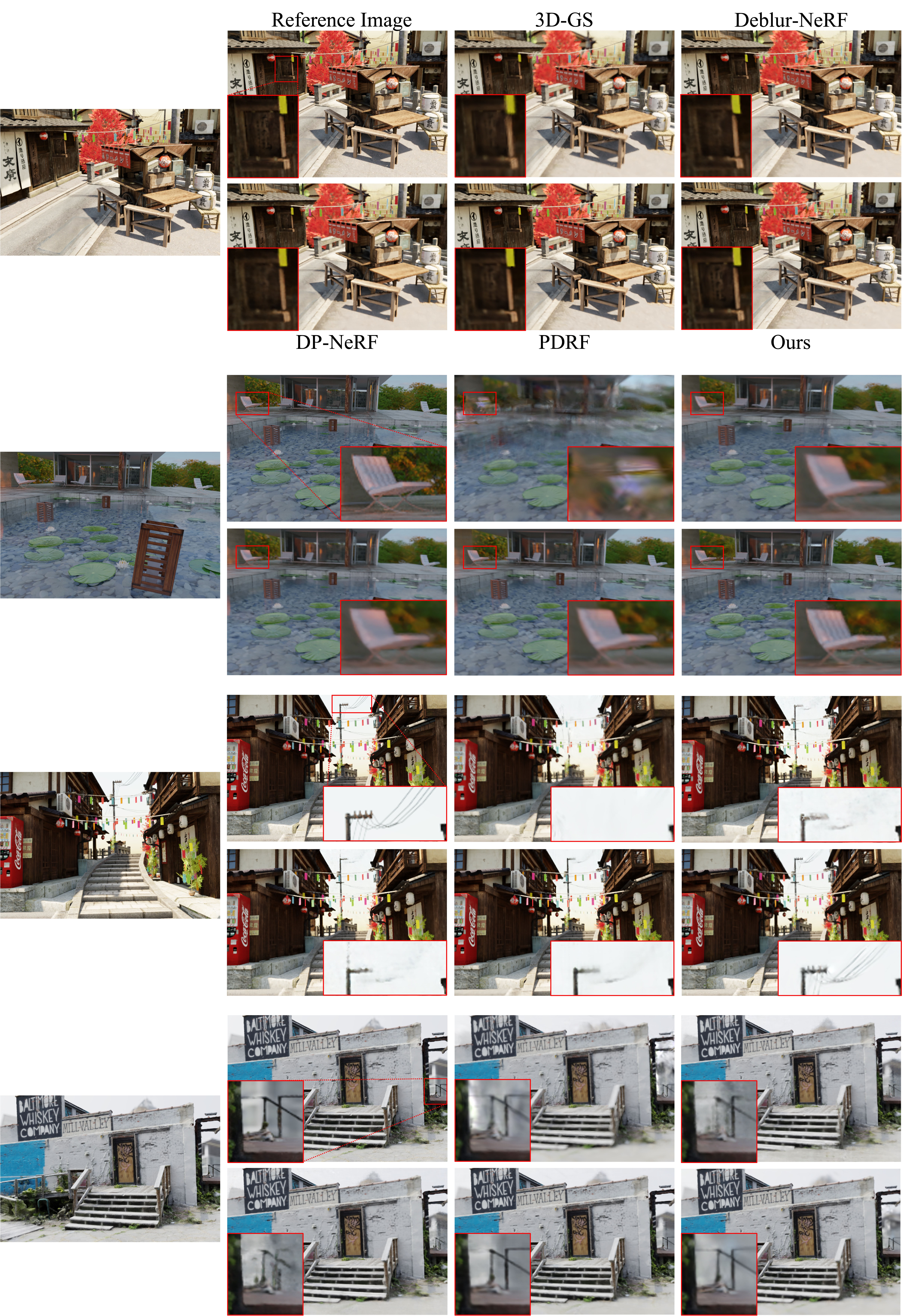}
    \caption{Qualitative results on synthetic defocus blur dataset.}
    \label{fig:qualitative_syn}
\end{figure*}

We further evaluate the proposed method under Learned Perceptual Image Patch Similarity (LPIPS) metric as shown in \cref{tab:real_defocus_lpips} and \cref{tab:synthetic_defocus_lpips} for real defocus blur and real camera motion blur dataset. Our method achieves state-of-the-art performance under LPIPS on both datasets. Also, we conducted experiments on the synthetic defocus blur (\cref{tab:quant_synthetic_defocus}, and~\cref{tab:lpips_combined}) and camera motion blur (\cref{tab:quant_synthetic_motion}, and~\cref{tab:lpips_combined}) dataset. For the synthetic defocus blur dataset, we set the learning rate for the position of 3D Gaussians to $4.8e-4$. The learning rate for scale and rotation attribute of 3D Gaussians is set to 0.015, and 0.005 respectively, $\lambda_s$ to 0.005, and $\lambda_p$ to 0.001 for the synthetic camera motion blur dataset. The rest hyperparameters are identical to those for real blur datasets. Qualitative results of each dataset are illustrated in \cref{fig:qualitative_real} to \cref{fig:qualitative_syn_motion}.
\begin{table}[]
    \centering
    \caption{Quantitative results on real defocus blur dataset tested under LPIPS metric. We color each cell as \colorbox{orange!50}{best} and \colorbox{yellow!50}{second best}.}
    \resizebox{\linewidth}{!}{
    \begin{tabular}{c|c|*{9}{c|}c}
    \hline
     & Cake & Caps & Cisco & Coral & Cupcake & Cups & Daisy & Sausage & Seal & Tools & Average \\
    \hline\hline
    NeRF~\cite{nerf} & 0.2250 & 0.2801 & 0.1256 & 0.2155 & 0.2689 & 0.2315 & 0.2621 & 0.2789 & 0.2680 & 0.1547 & 0.2310 \\
    3D-GS~\cite{kerbl20233d} & 0.2082 & 0.4329 & 0.1781 & 0.3111 & 0.2081 & 0.3211 & 0.2629 & 0.2855 & 0.3057 & 0.1953 & 0.2920 \\
    \hline
    Deblur-NeRF~\cite{ma2022deblur} & 0.1282 & 0.1612 & 0.0868 & \cellcolor{yellow!50}0.1160 & 0.1214 & 0.1271 & 0.1208 & 0.1796 & 0.1048 & 0.0610 & 0.1207 \\
    DP-NeRF~\cite{lee2022deblurred} & \cellcolor{yellow!50}0.1267 & \cellcolor{orange!50}0.1430 & \cellcolor{yellow!50}0.0840 & \cellcolor{orange!50}0.0960 & \cellcolor{yellow!50}0.1178 & \cellcolor{yellow!50}0.1035 & 0.1075 & \cellcolor{yellow!50}0.1473 & \cellcolor{yellow!50}0.1026 & \cellcolor{orange!50}0.0539 & \cellcolor{yellow!50}0.1082 \\
    PDRF-10~\cite{peng2022pdrf} & 0.1622 & 0.2854 & 0.0943 & 0.2335 & 0.1862 & 0.1370 & \cellcolor{yellow!50}0.1024 & 0.2126 & 0.1927 & 0.1395 & 0.1746 \\
    Ours & \cellcolor{orange!50}0.1158 & \cellcolor{yellow!50}0.1491 & \cellcolor{orange!50}0.0794 & 0.1310 &\cellcolor{orange!50}0.0985 & \cellcolor{orange!50}0.1028 &\cellcolor{orange!50}0.0949 &\cellcolor{orange!50}0.1405 & \cellcolor{orange!50}0.0980 & \cellcolor{yellow!50}0.0577 & \cellcolor{orange!50}0.1068 \\
    \hline
    \end{tabular}
    }
    \label{tab:real_defocus_lpips}
\end{table}

\begin{table}[]
    \centering
    \caption{Quantitative results on real camera motion blur dataset tested under LPIPS metric. We color each cell as \colorbox{orange!50}{best} and \colorbox{yellow!50}{second best}.}
    \resizebox{\linewidth}{!}{
    \begin{tabular}{c|c|*{9}{c|}c}
    \hline
     & Ball & Basket & Buick & Coffee & Decoration & Girl & Heron & Parterre & Puppet & Stair & Average \\
    \hline\hline
    NeRF~\cite{nerf} & 0.3992 & 0.3223 & 0.3502 & 0.2896 & 0.3633 & 0.3196 & 0.4129 & 0.4046 & 0.3389 & 0.4868 & 0.3687 \\
    3D-GS~\cite{kerbl20233d} & 0.3505 & 0.2925 & 0.2839 & 0.3113 & 0.3141 & 0.2843 & 0.3328 & 0.3197 & 0.2607 & 0.3888 & 0.3240 \\
    \hline
    Deblur-NeRF~\cite{ma2022deblur} & 0.2230 & 0.1481 & 0.1752 & 0.1244 & 0.1862 & 0.1687 & 0.2099 & 0.2161 & 0.1577 & 0.2102 & 0.1820 \\
    DP-NeRF~\cite{lee2022deblurred} & \cellcolor{yellow!50}{0.2088} & 0.1294 & \cellcolor{yellow!50}{0.1405} & \cellcolor{yellow!50}{0.1002} & \cellcolor{yellow!50}{0.1639} & \cellcolor{yellow!50}{0.1498} & \cellcolor{yellow!50}{0.1914} & \cellcolor{yellow!50}{0.1900} & \cellcolor{yellow!50}{0.1505} & \cellcolor{yellow!50}{0.1772} & \cellcolor{yellow!50}{0.1602} \\
    PDRF-10~\cite{peng2022pdrf} & 0.2487 & \cellcolor{yellow!50}{0.1241} & 0.1751 & 0.1424 & 0.2379 & 0.1828 & 0.2367 & 0.2639 & 0.1569 & 0.2430 & 0.2012 \\
    Ours & \cellcolor{orange!50}{0.1413} & \cellcolor{orange!50}{0.1155} & \cellcolor{orange!50}{0.0954} & \cellcolor{orange!50}{0.0676} & \cellcolor{orange!50}{0.0933} & \cellcolor{orange!50}{0.1011} & \cellcolor{orange!50}{0.1543} & \cellcolor{orange!50}{0.1206} & \cellcolor{orange!50}{0.0941} & \cellcolor{orange!50}{0.1123} & \cellcolor{orange!50}{0.1096} \\
    \hline
    \end{tabular}
    }
    \label{tab:synthetic_defocus_lpips}
\end{table}

\begin{table*}[]
    \centering
    \caption{Quantitative results on synthetic defocus blur dataset under PSNR and SSIM metrics. We color each cell as \colorbox{orange!50}{best} and \colorbox{yellow!50}{second best}.}
    \resizebox{\linewidth}{!}{
    \begin{tabular}{c|cc|cc|cc|cc|cc|cc|cc|c}
    \hline
         & \multicolumn{2}{c|}{Cozyroom}& \multicolumn{2}{c|}{Factory} &  \multicolumn{2}{c|}{Pool} & \multicolumn{2}{c|}{Tanabata} & \multicolumn{2}{c|}{Trolley}& \multicolumn{2}{c|}{Average} & FPS \\
     & PSNR$\uparrow$ & SSIM$\uparrow$ & PSNR$\uparrow$ & SSIM$\uparrow$ & PSNR$\uparrow$ & SSIM$\uparrow$ & PSNR$\uparrow$ & SSIM$\uparrow$ & PSNR$\uparrow$ & SSIM$\uparrow$ & PSNR$\uparrow$ & SSIM$\uparrow$ & $\uparrow$ \\
     \hline
     \hline
    NeRF~\cite{nerf} & 30.03 & 0.8926 & 25.36 & 0.7847 & 27.77 & 0.7266 & 23.90 & 0.7811 & 22.67 & 0.7103 & 25.93 & 0.7791 & $<$ 1 \\
    3D-GS~\cite{kerbl20233d} & 30.09  & 0.9024 & 24.52 & 0.8057 & 20.14 & 0.4451 & 23.08 & 0.7981 & 22.26 & 0.7400 & 24.02 & 0.7383 & \cellcolor{yellow!50}789  \\
    \hline
    Deblur-NeRF~\cite{ma2022deblur} & 31.85 & 0.9175 & 28.03 & 0.8628 & 30.52 & 0.8246 & 26.26 & 0.8517 & 25.18 & 0.8067 & 28.37 & 0.8527 & $<$ 1\\
    Sharp-NeRF~\cite{sharpnerf} & 31.32 & 0.9133 & 28.67 & 0.8979 & 30.51 & 0.8264 & 24.95 & 0.8536 & 26.03 & 0.8498 & 28.30 & 0.8682 & $<$ 1 \\
    DP-NeRF~\cite{lee2022deblurred} & \cellcolor{yellow!50}32.11 & 0.9215 & \cellcolor{yellow!50}29.26 & 0.8793 & \cellcolor{orange!50}31.44 & \cellcolor{yellow!50}0.8529 & 27.05 & 0.8635 & 26.79 & 0.8395 & 29.33 & 0.8713 & $<$ 1 \\
    PDRF-10~\cite{peng2022pdrf} & \cellcolor{orange!50}32.29 & \cellcolor{orange!50}0.9305 & \cellcolor{orange!50}30.90 & \cellcolor{orange!50}0.9138 & 30.97 &0.8408 & \cellcolor{orange!50}28.18 & \cellcolor{yellow!50}0.9006 & \cellcolor{orange!50}28.07 & \cellcolor{yellow!50}0.8799 & \cellcolor{orange!50}30.08 & \cellcolor{yellow!50}0.8931 & $<$ 1 \\
    Ours & 31.97 & \cellcolor{yellow!50}0.9275 & 29.16 & \cellcolor{yellow!50}0.9089 & \cellcolor{yellow!50}31.31 & \cellcolor{orange!50}0.8580 & \cellcolor{yellow!50}27.54 & \cellcolor{orange!50}0.9083 & \cellcolor{yellow!50}27.55 & \cellcolor{orange!50}0.8858 & \cellcolor{yellow!50}29.51  &\cellcolor{orange!50}0.8977 & \cellcolor{orange!50}798 \\
    \hline
    \end{tabular}
    }
    \label{tab:quant_synthetic_defocus}
\end{table*}

\begin{table*}[]
    \centering
    \caption{Quantitative results on synthetic camera motion blur dataset under PSNR and SSIM metrics. We color each cell as \colorbox{orange!50}{best} and \colorbox{yellow!50}{second best}.}
    \resizebox{\linewidth}{!}{
    \begin{tabular}{c|cc|cc|cc|cc|cc|cc|cc|c}
    \hline
         & \multicolumn{2}{c|}{Cozyroom}& \multicolumn{2}{c|}{Factory} &  \multicolumn{2}{c|}{Pool} & \multicolumn{2}{c|}{Tanabata} & \multicolumn{2}{c|}{Trolley}& \multicolumn{2}{c|}{Average} & FPS \\
     & PSNR$\uparrow$ & SSIM$\uparrow$ & PSNR$\uparrow$ & SSIM$\uparrow$ & PSNR$\uparrow$ & SSIM$\uparrow$ & PSNR$\uparrow$ & SSIM$\uparrow$ & PSNR$\uparrow$ & SSIM$\uparrow$ & PSNR$\uparrow$ & SSIM$\uparrow$ & $\uparrow$ \\
     \hline
     \hline
    NeRF~\cite{nerf} & 25.66 & 0.7941 & 19.32 & 0.4563 & 30.45 & 0.8354 & 22.22 & 0.6807 & 21.25 & 0.6370 & 23.78 & 0.6807 & $<$ 1 \\
    3D-GS~\cite{kerbl20233d} & 25.59 & 0.8076 & 18.11 & 0.4179 & 25.63 & 0.6326 & 21.35 & 0.6686 & 20.56 & 0.6257 & 22.25 & 0.6305 & \cellcolor{orange!50}1019 \\
    \hline
    Deblur-NeRF~\cite{ma2022deblur} & \cellcolor{yellow!50}32.08 & 0.9261 & 25.60 & 0.7750 &31.61 & 0.8682 & 27.11 & 0.8640 & 27.45 & 0.8632 & 28.77 & 0.8593 & $<$ 1 \\
    DP-NeRF~\cite{lee2022deblurred} & \cellcolor{orange!50}32.65 & \cellcolor{yellow!50}0.9317 & \cellcolor{yellow!50}25.91 & \cellcolor{yellow!50}0.7787 & \cellcolor{orange!50}31.96 & \cellcolor{yellow!50}0.8768 & \cellcolor{yellow!50}27.61 & 0.8748& \cellcolor{yellow!50}28.03 & \cellcolor{yellow!50}0.8752 & \cellcolor{yellow!50}29.23 & \cellcolor{yellow!50}0.8674 & $<$ 1 \\
    PDRF-10~\cite{peng2022pdrf} & 31.90 & \cellcolor{orange!50}0.9321 & \cellcolor{orange!50}26.56 & \cellcolor{orange!50}0.8102 & 31.29 & 0.8657 & \cellcolor{orange!50}28.21 & \cellcolor{orange!50}0.8952 & \cellcolor{orange!50}28.48 & \cellcolor{orange!50}0.8956 & \cellcolor{orange!50}29.29 & \cellcolor{orange!50}0.8798 & $<$ 1 \\
    Ours & 31.45 & 0.9222 & 24.01 & 0.7333 & \cellcolor{yellow!50}31.87 & \cellcolor{orange!50}0.8829 & 27.01 & \cellcolor{yellow!50}0.8807 & 26.88 & 0.8710 & 28.24 & 0.8580  & \cellcolor{yellow!50}919 \\
    \hline
    \end{tabular}
    }
    \label{tab:quant_synthetic_motion}
\end{table*}


\begin{table}[]
    \centering
    \caption{Quantiative results on synthetic camera motion blur dataset tested under LPIPS metric. We color each cell as \colorbox{orange!50}{best} and \colorbox{yellow!50}{second best}.}
    \resizebox{\linewidth}{!}{
    \begin{tabular}{c|c|c|c|c|c|c|c|c|c|c|c|c}
    \hline
             & \multicolumn{6}{c|}{Defocus blur}& \multicolumn{6}{c}{Camera motion blur} \\
               & Cozy2room & Factory & Pool & Tanabata & Wine & Average & Cozy2room & Factory & Pool & Tanabata & Wine & Average \\
    \hline\hline
    NeRF~\cite{nerf} & 0.0885 & 0.2351 & 0.3340 & 0.2142 & 0.2799 & 0.2303 & 0.2288 & 0.5304 & 0.1932 & 0.3653 & 0.3633 & 0.3362 \\
    3D-GS ~\cite{kerbl20233d} & 0.0692 & 0.1842 & 0.5094 & 0.1710 & 0.2281 & 0.2323 & 0.1645 & 0.4958 & 0.2632 & 0.3235 & 0.3390 & 0.3172 \\
    \hline
    Deblur-NeRF~\cite{ma2022deblur} & 0.0481 & 0.1127 & 0.1901 & 0.0995 & 0.1436 & 0.1188 & 0.0477 & 0.2687 & 0.1246 & 0.1228 & 0.1363 & 0.1400 \\
    DP-NeRF~\cite{lee2022deblurred} & \cellcolor{orange!50}{0.0386} & \cellcolor{orange!50}{0.1035} & \cellcolor{orange!50}{0.1563} & \cellcolor{orange!50}{0.0779} & \cellcolor{yellow!50}{0.1170} & \cellcolor{orange!50}{0.0987} &\cellcolor{orange!50}0.0355 & 0.2494 &\cellcolor{yellow!50}0.0908 & 0.1033 & 0.1129 & 0.1184 \\
    PDRF-10~\cite{peng2022pdrf} & 0.0518 & \cellcolor{yellow!50}0.1066 & 0.1893 & 0.0819 & 0.1210 & 0.1101 & 0.0448 &\cellcolor{orange!50}0.1499 & 0.1345 &\cellcolor{yellow!50}0.1025 &\cellcolor{orange!50}0.0939 &\cellcolor{orange!50}0.1051 \\
    Ours & \cellcolor{yellow!50}0.0424 & 0.1079 & \cellcolor{yellow!50}0.1587 & \cellcolor{yellow!50}0.0821 & \cellcolor{orange!50}0.1163 & \cellcolor{yellow!50}0.1015 & \cellcolor{yellow!50}0.0367 & \cellcolor{yellow!50}0.2326 & \cellcolor{orange!50}0.0751 & \cellcolor{orange!50}0.0785 & \cellcolor{yellow!50}0.1028 & \cellcolor{orange!50}0.1051\\
    \hline
    \end{tabular}
    }
    \label{tab:lpips_combined}
\end{table}


\subsubsection{Depth-based pruning} 
\begin{figure}[]
\begin{center}
\includegraphics[width=0.9\linewidth]{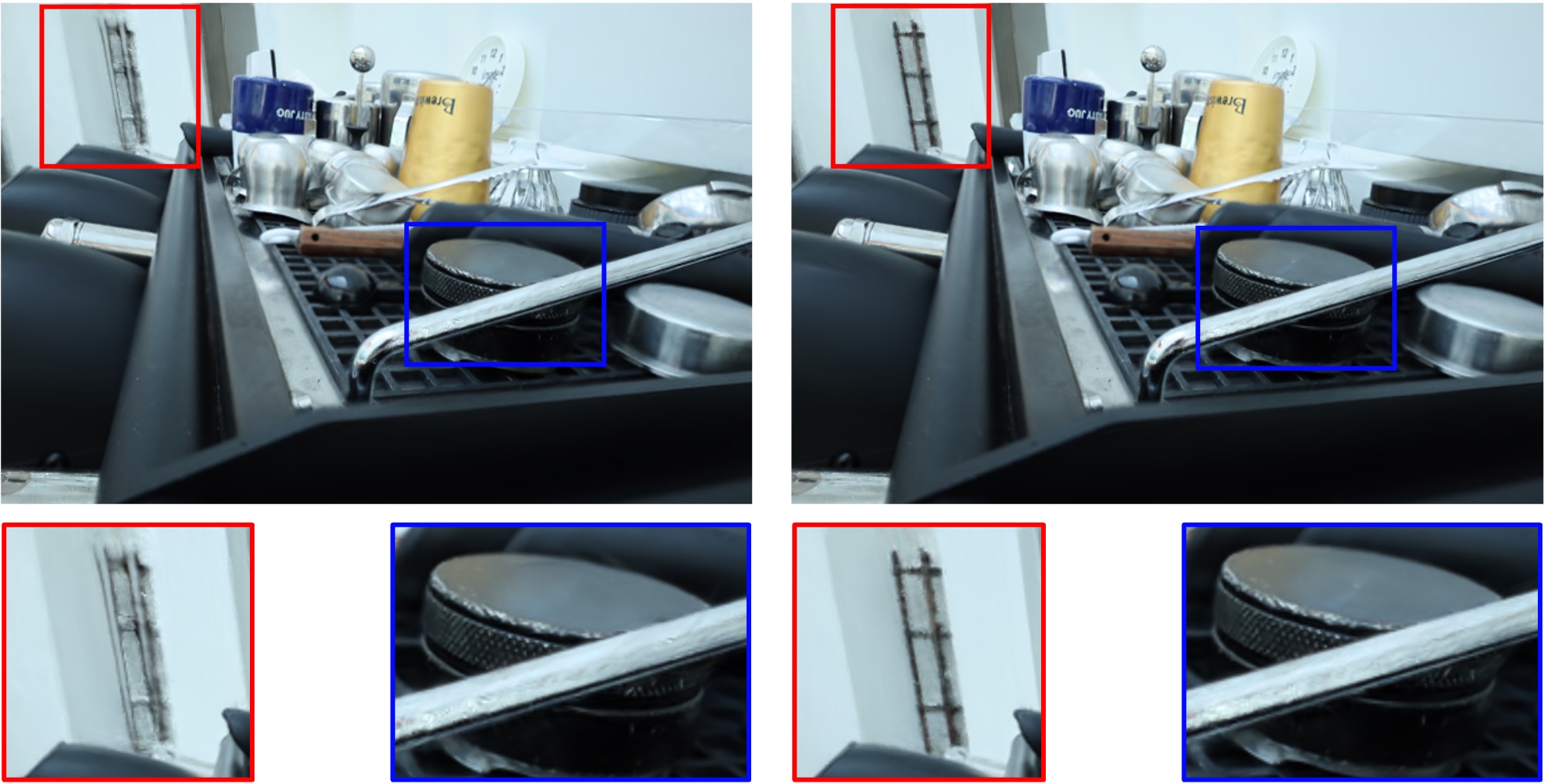}
\end{center}
\vspace{-0.5cm}
   \caption{Comparison on applying depth-based pruning. Top row: Rendered image from the model with depth-dependent pruning. Bottom row: Rendered image from the model with naive pruning as 3D-GS does.}
\label{fig:depth_pruning}
\end{figure}

We conduct an ablation study on depth-based pruning. To address excessive sparsity of point cloud at the far plane, we prune the points on the far plane less to maintain more numbers of points. \cref{tab:depth} shows our depth-based pruning can preserve more points located at the far plane which leads to better reconstruction quality than naive pruning, which prunes the points with a single threshold regardless of the positions of the points. In addition, \cref{fig:depth_pruning} shows a failure to reconstruct objects at the far plane when naive pruning is used, while objects lying on the near-end of the scene are well reconstructed. Meanwhile, ours, with depth-based pruning, can render clean objects on both near and far planes.

\begin{table}[h]
    \centering
    \caption{Ablation study on depth-depending pruning. Naive pruning stands for using naive points pruning from 3D-GS and Depth-based pruning stands for applying our depth-based pruning.}
    \begin{tabular}{c|c|c}
    \hline
    Methods & PSNR $\uparrow$ & SSIM $\uparrow$ \\
    \hline\hline
    Naive pruning & 23.52 & 0.7394 \\
    Depth-based pruning & 23.73 & 0.7474 \\
    \hline
    \end{tabular}
    \label{tab:depth}
    \vspace{-0.5cm}
\end{table}

\subsubsection{Ablation study on $M$}
We run an ablation study on the hyperparameter $M$, the number of the moments to be averaged. As shown in \cref{tab:ablation_M} our method shows similar performance when $ 5 \leq M \leq 10$. Considering the increasing training time with higher $M$, we set $M$ to 5.

\begin{table}[h!]
    \centering
    \caption{Ablation on $M$ under real camera motion blur dataset.}
    \begin{tabular}{c|c|c|c|c|c}
    \hline
    & PSNR & Training time & & PSNR & Training time \\
    \hline\hline
    $M = 2$ & 24.81 & $\approx$ 8.0 mins & $M = 10$ & 26.62 & $\approx$ 22.4 mins \\
    $M = 5$ & 26.66 & $\approx$ 13.4 mins & $M = 20$ & 25.81 & $\approx$ 43.4 mins \\
    \hline
    \end{tabular}
    \label{tab:ablation_M}
\end{table}

\subsubsection{Ablation study on extra points allocation}
In this section, we evaluate the effect of adding extra points to the sparse point cloud. As shown in \cref{fig:extra_points}, directly using sparse point cloud without any point densification only represents the objects with a small number of points or fails to model the tiny objects. Meanwhile, in case extra points are added to the point cloud, points successfully represent the objects densely. Also, the quantitative result is presented in \cref{tab:extra_points}. It shows assigning valid color features to the additional points is important to deblur the scene and reconstruct the fine details.

\begin{table}[]
    \centering
    \caption{Ablation study on adding the extra points. w/ Random Colors stands for uniformly allocating points to the point cloud but color features are randomly initialized, rather than interpolated from neighboring points.}
    \begin{tabular}{c|c|c}
    \hline
    Methods & PSNR $\uparrow$ & SSIM $\uparrow$ \\
    \hline\hline
    w/o Extra Points & 23.34 & 0.7374 \\
    w/ Random Colors & 23.46 & 0.7411 \\
    w/ Extra Points & 23.73 & 0.7474 \\
    \hline
    \end{tabular}
    \label{tab:extra_points}
\end{table}

\subsubsection{Training time comparison}
\cref{tab:training_time} shows training time and the number of the Gaussians and \cref{tab:moudle_time} describe the running time MLP and rasterization. 
Although MLP introduces additional costs, our approach is more efficient than prior works that generate kernels and apply convolutions. 3D-GS hardly models the camera motion blur thus fewer 3D Gaussians are involved, dropping the training time. Also, the proposed method uses $M=5$, a relatively smaller number of frames compared to the existing works, leading to faster training time.

\begin{table}[h!]
    \centering
    \caption{Comparison on training time (RTX 4090 GPU).}
    \begin{tabular}{c|cc|cc}
    \hline
         & \multicolumn{2}{c|}{Camera Motion Blur}& \multicolumn{2}{c}{Defocus Blur}\\
     & Time (min) & \# Gaussians & Time (min) & \# Gaussians \\
     \hline
     \hline
    3D-GS & $\approx$ 2.1 & 63,549 & $\approx$ 3.0 & 308,986\\
    Ours & $\approx$ 13.4 & 139,545 & $\approx$ 8.2 & 317,964 \\
    \hline
    \end{tabular}
    \label{tab:training_time}
\end{table}

\begin{table}[h!]
    \centering
    \caption{Running time of each module.}
    \begin{tabular}{c|c}
    \hline
        MLP forwarding 1.33 ms & Rasterization 1.00 ms \\
    \hline
    \end{tabular}
    \label{tab:moudle_time}
\end{table}

\subsubsection{Images from nearly the same viewpoints} 
\begin{figure}
\centering
\includegraphics[width=\textwidth]{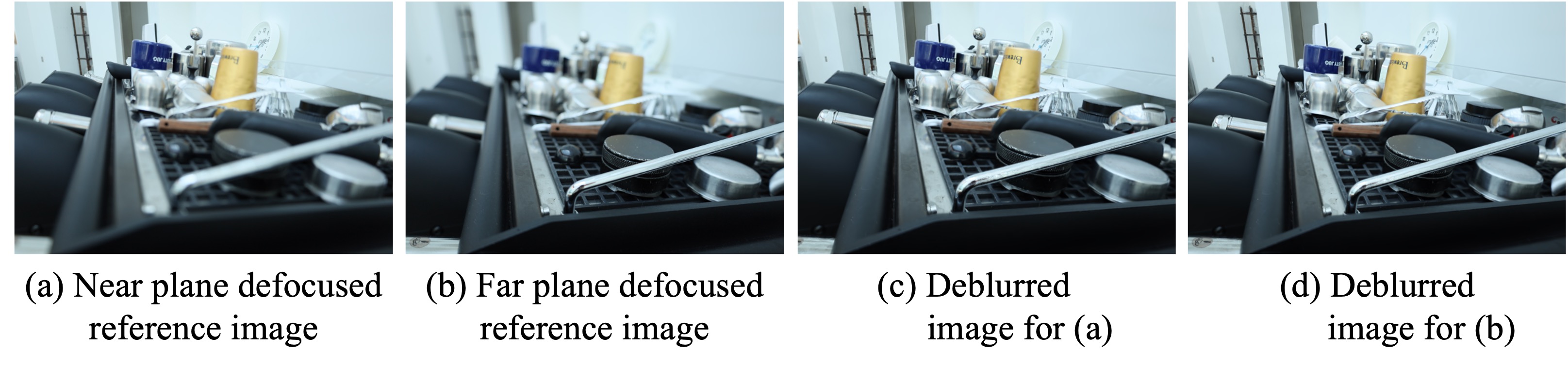}
   \caption{Deblurred images at nearly the same viewpoints.}
\label{fig:closeview}
\end{figure}

Two training images in \cref{fig:closeview} are taken with nearly the same viewpoints but (a) is defocused at the near plane while (b) shows defocus blur at the far plane. Even if they are captured at very close views, the proposed method can deblur only the blurry regions, keeping the clean regions as they are, which highlights its view-dependent functionality.

\begin{figure*}[p]
    \centering
    \includegraphics[width=\textwidth, keepaspectratio]{figs/qualitative_synthetic_motin.jpg}
    \caption{Qualitative results on synthetic camera motion blur dataset.}
    \label{fig:qualitative_syn_motion}
\end{figure*}

\end{document}